\documentclass[runningheads]{llncs}
\usepackage{graphicx}
\usepackage{amsmath,amssymb} 
\usepackage{color}
\usepackage[width=122mm,left=12mm,paperwidth=146mm,height=193mm,top=12mm,paperheight=217mm]{geometry}

\usepackage{xspace}
\usepackage[dvipsnames]{xcolor}
\usepackage{graphicx}
\usepackage[flushleft]{threeparttable}
\usepackage{stackengine}
\usepackage{makecell}
\usepackage{enumitem}

\newcommand{\vpara}[1]{\vspace{0.05in}\noindent\textbf{#1}}
\newcommand{\oneshot}[2]{\noindent\textbf{Proposition {#1}.\ }{#2}}

\newcommand*\lstackon[3][\stackgap]{\stackengine%
	{#1}{#2}{#3}{O}{\stackalignment}{\quietstack}{\useanchorwidth}{L}}
\newcommand*\lstackunder[3][\stackgap]{\stackengine%
	{#1}{#2}{#3}{U}{\stackalignment}{\quietstack}{\useanchorwidth}{L}}
\DeclareRobustCommand\onedot{\futurelet\@let@token\@onedot}
\def\@onedot{\ifx\@let@token.\else.\null\fi\xspace}

\def\etal{\emph{et al.}\xspace}
\newif\ifsubmit
\submitfalse
\ifsubmit
\newcommand{\my}[1]{}
\newcommand{\revmy}[1]{}
\newcommand{\JK}[1]{}
\else
\newcommand{\my}[1]{{\bf \textcolor{magenta}{MY: #1}}}
\newcommand{\revmy}[1]{\textcolor{magenta}{#1}}

\newcommand{\JK}[1]{{\bf \textcolor{red}{[Jan: #1]}}}
\fi
\newcommand{\ignore}[1]{}

\usepackage{colortbl}
\usepackage{arydshln,graphicx,xcolor,array}
\usepackage[implicit=false,hidelinks]{hyperref}

\begin{document}
	\pagestyle{headings}
	\mainmatter
	\def\ECCV18SubNumber{143}  
	
	\title{Multimodal Unsupervised \\Image-to-Image Translation} 

\titlerunning{Multimodal Unsupervised Image-to-Image Translation}
%

	\authorrunning{\hfill Xun Huang, Ming-Yu Liu, Serge Belongie, Jan Kautz}
	\author{Xun Huang$^{1}$, Ming-Yu Liu$^{2}$, Serge Belongie$^{1}$, Jan Kautz$^{2}$}
%
%

	\institute{Cornell University$^{1}$\hspace{0.5in}NVIDIA$^{2}$}
	
\maketitle              

	\begin{abstract}
		Unsupervised image-to-image translation is an important and challenging problem in computer vision.
		Given an image in the source domain, the goal is to learn the conditional distribution of corresponding images in the target domain, without seeing any examples of corresponding image pairs.
		While this conditional distribution is inherently multimodal, existing approaches make an overly simplified assumption, modeling it as a deterministic one-to-one mapping. As a result, they fail to generate diverse outputs from a given source domain image. To address this limitation, we propose a Multimodal Unsupervised Image-to-image \mbox{Translation~(MUNIT)} framework. We assume that the image representation can be decomposed into a content code that is domain-invariant, and a style code that captures domain-specific properties. To translate an image to another domain, we recombine its content code with a random style code sampled from the style space of the target domain. We analyze the proposed framework and establish several theoretical results. Extensive experiments with comparisons to state-of-the-art approaches further demonstrate the advantage of the proposed framework. Moreover, our framework allows users to control the style of translation outputs by providing an example style image.
		Code and pretrained models are available at \href{https://github.com/nvlabs/MUNIT}{https://github.com/nvlabs/MUNIT}.
		

	\end{abstract}
	\section{Introduction}
	Many problems in computer vision aim at translating images from one domain to another, including
	super-resolution~\cite{dong2014learning}, colorization~\cite{zhang2016colorful}, inpainting~\cite{pathak2016context}, attribute transfer~\cite{laffont2014transient}, and style transfer~\cite{gatys2016image}. This cross-domain image-to-image translation setting has therefore received significant attention~\cite{isola2017image,yi2017dualgan,zhu2017unpaired,kim2017learning,taigman2017unsupervised,zhu2017toward,liu2016coupled,chen2017photographic,liang2017generative,liu2017unsupervised,benaim2017one,royer2017xgan,gan2017triangle,choi2017stargan,wang2018high,shrivastava2017learning,bousmalis2017unsupervised,wolf2017unsupervised,tau2018role,hoshen2018identifying}. When the dataset contains paired examples, this problem can be approached by a conditional generative model~\cite{isola2017image} or a simple regression model~\cite{chen2017photographic}.  In this work, we focus on the much more challenging setting when such supervision is unavailable.
	
	In many scenarios, the cross-domain mapping of interest is multimodal. For example, a winter scene could have many possible appearances during summer due to weather, timing, lighting, etc. Unfortunately, existing techniques usually assume a deterministic~\cite{zhu2017unpaired,kim2017learning,taigman2017unsupervised} or unimodal~\cite{liu2017unsupervised} mapping. As a result, they fail to capture the full distribution of possible outputs. Even if the model is made stochastic by injecting noise, the network usually learns to ignore it~\cite{isola2017image,mathieu2016deep}.

\begin{figure*}[!tb]
		\definecolor{myred}{HTML}{C00000}
		\definecolor{myblue}{HTML}{0070C0}
		\centering
		\includegraphics[width=1\textwidth]{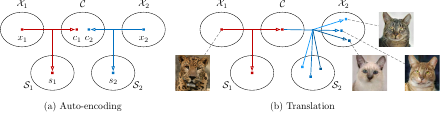}
		\caption{An illustration of our method. (a) Images in each domain $\mathcal{X}_{i}$ are encoded to a shared content space $\mathcal{C}$ and a domain-specific style space $\mathcal{S}_{i}$. Each encoder has an inverse decoder omitted from this figure. (b) To translate an image in $\mathcal{X}_{1}$ (\textit{e.g.}, a leopard) to $\mathcal{X}_{2}$ (\textit{e.g.}, domestic cats), we recombine the content code of the input with a random style code in the target style space. \mbox{Different style codes lead to different outputs.}}
		\label{fig:illustration}
	\end{figure*}

	In this paper, we propose a principled framework for the Multimodal UNsupervised Image-to-image Translation~(MUNIT) problem. As shown in Fig.~\ref{fig:illustration}~(a), our framework makes several assumptions. We first assume that the latent space of images can be decomposed into a content space and a style space. We further assume that images in different domains share a common content space but not the style space.	To translate an image to the target domain, we recombine its content code with a random style code in the target style space (Fig.~\ref{fig:illustration}~(b)). The content code encodes the information that should be preserved during translation, while the style code represents remaining variations that are not contained in the input image. By sampling different style codes, our model is able to produce diverse and multimodal outputs. Extensive experiments demonstrate the effectiveness of our method in modeling multimodal output distributions and its superior image quality compared with state-of-the-art approaches. Moreover, the decomposition of content and style spaces allows our framework to perform example-guided image translation, in which the style of the translation outputs are controlled by a user-provided example image in the target domain.

	\section{Related Works}	\label{sec::related}

	\vpara{Generative adversarial networks~(GANs).} The GAN framework~\cite{goodfellow2014generative} has achieved impressive results in image generation. In GAN training, a generator is trained to fool a discriminator which in turn tries to distinguish between generated samples and real samples.
	Various improvements to GANs have been proposed, such as multi-stage generation~\cite{denton2015deep,wang2016generative,yang2017lrgan,huang2017sgan,zhang2017stackgan,karras2018progressive}, better training objectives~\cite{salimans2016improved,zhao2016energy,arjovsky2017wasserstein,berthelot2017began,mao2017least,tolstikhin2018wasserstein}, and combination with auto-encoders~\cite{larsen2016autoencoding,dosovitskiy2016generating,rosca2017variational,li2017alice,srivastava2017veegan}. 
	In this work, we employ GANs to align the distribution of translated images with real images \mbox{in the target domain.}
	
	\vpara{Image-to-image translation.} Isola~\etal~\cite{isola2017image} propose the first unified framework for image-to-image translation based on conditional GANs, which has been extended to generating high-resolution images by Wang~\etal~\cite{wang2018high}. 
	Recent studies have also attempted to learn image translation without supervision. 
	This problem is inherently ill-posed and requires additional constraints.
	Some works enforce the translation to preserve certain properties of the source domain data, such as pixel values~\cite{shrivastava2017learning}, pixel gradients~\cite{bousmalis2017unsupervised}, semantic features~\cite{taigman2017unsupervised}, class labels \cite{bousmalis2017unsupervised}, or pairwise sample distances~\cite{benaim2017one}. Another popular constraint is the cycle consistency loss~\cite{yi2017dualgan,zhu2017unpaired,kim2017learning}. It enforces that if we translate an image to the target domain and back, we should obtain the original image. In addition, Liu \etal~\cite{liu2017unsupervised} propose the UNIT framework, which assumes a shared latent space such that corresponding images in two domains are mapped to the same latent code. 
	
	A significant limitation of most existing image-to-image translation methods is the lack of diversity in the translated outputs. 
	To tackle this problem, some works propose to simultaneously generate multiple outputs given the same input and encourage them to be different~\cite{chen2017photographic,ghosh2017multi,bansal2018pixelnn}. Still, these methods can only generate a discrete number of outputs. Zhu \etal~\cite{zhu2017toward} propose a BicycleGAN that can model continuous and multimodal distributions. However, all the aforementioned methods require pair supervision, while our method does not. A couple of concurrent works also recognize this limitation and propose extensions of CycleGAN/UNIT for multimodal mapping~\cite{almahairi2018augmented}/\cite{lee2018diverse}.
	
	Our problem has some connections with multi-domain image-to-image translation~\cite{choi2017stargan,anoosheh2017combogan,hui2017unsupervised}.
	Specifically, when we know how many modes each domain has and the mode each sample belongs to, it is possible to treat each mode as a separate domain and use multi-domain image-to-image translation techniques to learn a mapping between each pair of modes, thus achieving multimodal translation. However, in general we do not assume such information is available. Also, our stochastic model can represent continuous output distributions, while \cite{choi2017stargan,anoosheh2017combogan,hui2017unsupervised} still use a deterministic model for each pair of domains.
	
	\vpara{Style transfer.}  Style transfer aims at modifying the style of an image while preserving its content, which is closely related to image-to-image translation. Here, we make a distinction between example-guided style transfer, in which the target style comes from a single example, and collection style transfer, in which the target style is defined by a collection of images. Classical style transfer approaches~\cite{gatys2016image,hertzmann2001image,li2016combining,johnson2016perceptual,huang2017adain,li2017universal,li2018closed} typically tackle the former problem, whereas image-to-image translation methods have been demonstrated to perform well in the latter~\cite{zhu2017unpaired}. We will show that our model is able to address both problems, thanks to its disentangled representation of content and style.
	
	\vpara{Learning disentangled representations.} Our work draws inspiration from recent works on disentangled representation learning. For example, InfoGAN~\cite{chen2016infogan} and $\beta$-VAE~\cite{higgins2017beta} have been proposed to learn disentangled representations without supervision. Some other works~\cite{tenenbaum1997separating,bousmalis2016domain,villegas2017decomposing,mathieu2016disentangling,denton2017unsupervised,tulyakov2018mocogan,donahue2018semantically,shen2017style} focus on disentangling content from style. Although it is difficult to define content/style and different works use different definitions, we refer to ``content'' as the underling spatial structure and ``style'' as the rendering of the structure. In our setting, we have two domains that share the same content distribution but have different style distributions.

	\section{Multimodal Unsupervised Image-to-image Translation}
	\label{sec:framework}
	

	\subsection{Assumptions}
	\label{sec:assumptions}
	
	Let $x_{1}\in\mathcal{X}_{1}$ and $x_{2}\in\mathcal{X}_{2}$ be images from two different image domains. In the unsupervised image-to-image translation setting, we are given samples drawn from two marginal distributions $p(x_{1})$ and $p(x_{2})$, without access to the joint distribution $p(x_{1},x_{2})$. Our goal is to estimate the two conditionals $p(x_{2}|x_{1})$ and $p(x_{1}|x_{2})$ with learned image-to-image translation models $p(x_{1\rightarrow 2}|x_{1})$ and $p(x_{2\rightarrow 1}|x_{2})$, where $x_{1\rightarrow 2}$ is a sample produced by translating $x_{1}$ to $\mathcal{X}_{2}$~(similar for $x_{2\rightarrow 1}$). In general, $p(x_{2}|x_{1})$ and $p(x_{1}|x_{2})$ are complex and multimodal distributions, in which case a deterministic translation model does not work well.
	
	To tackle this problem, we make a \textit{partially shared latent space assumption}. Specifically, we assume that each image $x_{i}\in\mathcal{X}_{i}$ is generated from a content latent code $c\in \mathcal{C}$ that is shared by both domains, and a style latent code $s_{i}\in \mathcal{S}_{i}$ that is specific to the individual domain. In other words, a pair of corresponding images $(x_{1},x_{2})$ from the joint distribution is generated by $x_{1} = G^{*}_{1}(c, s_{1})$ and $x_{2} = G^{*}_{2}(c, s_{2})$, where $c, s_{1}, s_{2}$ are from some prior distributions and $G^{*}_{1}$, $G^{*}_{2}$ are the underlying generators. We further assume that $G^{*}_{1}$ and $G^{*}_{2}$ are deterministic functions and have their inverse encoders $E^{*}_{1}=(G^{*}_{1})^{-1}$ and $E^{*}_{2}=(G^{*}_{2})^{-1}$. 
	Our goal is to learn the underlying generator and encoder functions with neural networks. Note that although the encoders and decoders are deterministic, $p(x_{2}|x_{1})$ is a continuous distribution due to the \mbox{dependency of $s_{2}$}.

	Our assumption is closely related to the shared latent space assumption proposed in UNIT~\cite{liu2017unsupervised}. While UNIT assumes a fully shared latent space, we postulate that only part of the latent space~(the content) can be shared across domains whereas the other part~(the style) is domain specific, which is a more reasonable assumption when the cross-domain mapping is \emph{many-to-many}. 

	\subsection{Model}
	\label{sec:model}	
\begin{figure*}[!tb]
		\definecolor{myred}{HTML}{C00000}
		\definecolor{myblue}{HTML}{0070C0}
		\centering
		\includegraphics[width=0.9\textwidth]{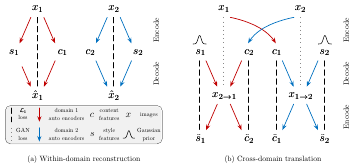}
		\caption{Model overview. Our image-to-image translation model consists of two auto-encoders~(denoted by \textcolor{myred}{red} and \textcolor{myblue}{blue} arrows respectively), one for each domain. The latent code of each auto-encoder is composed of a content code $c$ and a style code $s$. We train the model with adversarial objectives~(dotted lines) that ensure the translated images to be indistinguishable from real images in the target domain, as well as bidirectional reconstruction objectives~(dashed lines) that reconstruct both images and latent codes. 
		} 		
		\label{fig:overview}
	\end{figure*}

	Fig.~\ref{fig:overview} shows an overview of our model and its learning process. Similar to~Liu \etal\cite{liu2017unsupervised}, our translation model consists of an encoder $E_{i}$ and a decoder $G_{i}$  for each domain $\mathcal{X}_{i}$ ($i=1,2$). As shown in Fig.~\ref{fig:overview}~(a), the latent code of each auto-encoder is factorized into a content code $c_{i}$ and a style code $s_{i}$, where $(c_{i}, s_{i}) = (E_{i}^{c}(x_{i}), E_{i}^{s}(x_{i})) = E_{i}(x_{i})$. Image-to-image translation is performed by swapping encoder-decoder pairs, as illustrated in Fig.~\ref{fig:overview}~(b). For example, to translate an image $x_{1}\in \mathcal{X}_{1}$ to $\mathcal{X}_{2}$, we first extract its content latent code $c_{1} = E^{c}_{1}(x_{1})$ and randomly draw a style latent code $s_{2}$ from the prior distribution $q(s_{2})\sim \mathcal{N}(0, \mathbf{I})$. We then use $G_{2}$ to produce the final output image $x_{1\rightarrow 2} = G_{2}(c_{1}, s_{2})$. We note that although the prior distribution is unimodal, the output image distribution can be multimodal thanks to the nonlinearity of the decoder.
	
	Our loss function comprises a \textit{bidirectional reconstruction loss} that ensures the encoders and decoders are inverses, and an \textit{adversarial loss} that matches the distribution of translated images to the image distribution in the target domain.
	
	\vpara{Bidirectional reconstruction loss.} To learn pairs of encoder and decoder that are inverses of each other, we use objective functions that encourage reconstruction in both image \mbox{$\rightarrow$ latent $\rightarrow$ image and latent $\rightarrow$ image $\rightarrow$ latent directions}:
	\begin{itemize}[leftmargin=*]
		\item {\bf Image reconstruction.} Given an image sampled from the data distribution, we should be able to reconstruct it after encoding and decoding.
		\begin{align}
		\mathcal{L}^{x_{1}}_{\text{recon}} =\mathbb{E}_{x_{1}\sim p(x_{1})}[||G_{1}(E_{1}^{c}(x_{1}), E_{1}^{s}(x_{1}))-x_{1}||_{1}]
		\end{align}
		\item {\bf Latent reconstruction.} Given a latent code (style and content) sampled from the latent distribution at translation time, we should be able to reconstruct it after decoding and encoding.
		\begin{align}
		\mathcal{L}^{c_{1}}_{\text{recon}} = \mathbb{E}_{c_{1}\sim p(c_{1}), s_{2}\sim q(s_{2})}[||E^{c}_{2}(G_{2}(c_{1},s_{2}))-c_{1}||_{1}]\\
		\mathcal{L}^{s_{2}}_{\text{recon}} = \mathbb{E}_{c_{1}\sim p(c_{1}), s_{2}\sim q(s_{2})}[||E^{s}_{2}(G_{2}(c_{1},s_{2}))-s_{2}||_{1}]
		\end{align}
		where $q(s_{2})$ is the prior $\mathcal{N}(0, \mathbf{I})$, \mbox{$p(c_{1})$ is given by $c_{1} = E^{c}_{1}(x_{1})$ and $x_{1}\sim p(x_{1})$.}
	\end{itemize}
	We note the other loss terms $\mathcal{L}^{x_{2}}_{\text{recon}}$, $\mathcal{L}^{c_{2}}_{\text{recon}}$, and $\mathcal{L}^{s_{1}}_{\text{recon}}$ are defined in a similar manner. We use $\mathcal{L}_{1}$ reconstruction loss as it encourages sharp output images.
	
	The style reconstruction loss $\mathcal{L}^{s_{i}}_{\text{recon}}$ is reminiscent of the latent reconstruction loss used in the prior works~\cite{zhu2017toward,huang2017sgan,srivastava2017veegan,chen2016infogan}. It has the effect on encouraging diverse outputs given different style codes. The content reconstruction loss $\mathcal{L}^{c_{i}}_{\text{recon}}$ encourages the translated image to preserve semantic \mbox{content of the input image.}

	\vpara{Adversarial loss.} We employ GANs to match the distribution of translated images to the target data distribution. In other words, images generated by our model should be indistinguishable from real images \mbox{in the target domain.}
	\begin{align}
	\mathcal{L}^{x_{2}}_{\text{GAN}}&=\mathbb{E}_{c_{1}\sim p(c_{1}), s_{2}\sim q(s_{2})}[\log(1-D_{2}(G_{2}(c_{1},s_{2})))] + \mathbb{E}_{x_{2}\sim p(x_{2})}[\log D_{2}(x_{2})] 
	\end{align}
	where $D_{2}$ is a discriminator that tries to distinguish between translated images and real images in $\mathcal{X}_{2}$. The discriminator $D_{1}$ and loss $\mathcal{L}^{x_{1}}_{\text{GAN}}$ are defined similarly.
	
	\vpara{Total loss.} We jointly train the encoders, decoders, and discriminators to optimize the final objective, which is a weighted sum of the adversarial loss and the bidirectional reconstruction loss terms.
	\begin{align} \underset{E_{1}, E_{2}, G_{1}, G_{2}}\min\ \underset{D_{1}, D_{2}}\max\ \mathcal{L}(E_{1}, E_{2}, G_{1}, G_{2}, D_{1}, D_{2}) = \mathcal{L}^{x_{1}}_{\text{GAN}} + \mathcal{L}^{x_{2}}_{\text{GAN}}\ + \notag \\ 
	\lambda_{x}(\mathcal{L}^{x_{1}}_{\text{recon}}+\mathcal{L}^{x_{2}}_{\text{recon}})+\lambda_{c}(\mathcal{L}^{c_{1}}_{\text{recon}}+\mathcal{L}^{c_{2}}_{\text{recon}})+\lambda_{s}(\mathcal{L}^{s_{1}}_{\text{recon}}+\mathcal{L}^{s_{2}}_{\text{recon}})
	\label{equ:loss}
	\end{align}
	where $\lambda_{x}$, $\lambda_{c}$, $\lambda_{s}$ are weights that control the importance of reconstruction terms.
	
	\section{Theoretical Analysis}
	\label{sec:theory}
	We now establish some theoretical properties of our framework. Specifically, we show that minimizing the proposed loss function leads to 1) matching of latent distributions during encoding and generation, 2) matching of two joint image distributions induced by our framework, and 3) enforcing a weak form of cycle consistency constraint. \mbox{All the proofs are given in Appendix~\ref{app:proofs}.}
	
	First, we note that the total loss in Eq.~(\ref{equ:loss}) is minimized when the translated distribution matches the data distribution and the encoder-decoder are inverses.
	\begin{proposition}
		Suppose there exists $E^{*}_{1}$, $E^{*}_{2}$, $G^{*}_{1}$, $G^{*}_{2}$ such that: 1) $E^{*}_{1} = (G^{*}_{1})^{-1}$ and $E^{*}_{2} = (G^{*}_{2})^{-1}$, and 2) $p(x_{1\rightarrow 2}) = p(x_{2})$ and $p(x_{2\rightarrow 1}) = p(x_{1})$. Then $E^{*}_{1}$, $E^{*}_{2}$, $G^{*}_{1}$, $G^{*}_{2}$ minimizes $\mathcal{L}(E_{1}, E_{2}, G_{1}, G_{2})=\underset{D_{1}, D_{2}}\max\ \mathcal{L}(E_{1}, E_{2}, G_{1}, G_{2}, D_{1}, D_{2})$ (Eq.~(\ref{equ:loss})).
	\end{proposition}

	\subsubsection{Latent Distribution Matching} For image generation, existing works on combining auto-encoders and GANs need to match the encoded latent distribution with the latent distribution the decoder receives at generation time, using either KLD loss~\cite{liu2017unsupervised,larsen2016autoencoding} or adversarial loss~\cite{royer2017xgan,rosca2017variational} in the latent space. The auto-encoder training would not help GAN training if the decoder received a very different latent distribution during generation. Although our loss function does not contain terms that explicitly encourage the match of latent distributions, \mbox{it has the effect of matching them implicitly.}
	\begin{proposition}
		When optimality is reached, we have:
		\label{proposition:2}
		$$p(c_{1})=p(c_{2}),\ p(s_{1})=q(s_{1}),\ p(s_{2})=q(s_{2})$$		
	\end{proposition}
	The above proposition shows that at optimality, the encoded style distributions match their Gaussian priors. Also, the encoded content distribution matches the distribution at generation time, which is just the encoded distribution from the other domain. This suggests that the content space becomes domain-invariant.

	\subsubsection{Joint Distribution Matching} Our model learns two conditional distributions $p(x_{1\rightarrow 2}| x_{1})$ and $p(x_{2\rightarrow 1}| x_{2})$, which, together with the data distributions, define two joint distributions $p(x_{1}, x_{1\rightarrow 2})$ and  $p(x_{2\rightarrow 1}, x_{2})$. Since both of them are designed to approximate the same underlying joint distribution $p(x_{1}, x_{2})$, it is desirable that they are consistent with each other, \textit{i.e.}, $p(x_{1}, x_{1\rightarrow 2}) = p(x_{2\rightarrow 1}, x_{2})$.
	
	Joint distribution matching provides an important constraint for unsupervised image-to-image translation and is behind the success of many recent methods. Here, we show our model matches the joint distributions at optimality.
	\begin{proposition}
		When optimality is reached,  we have
		$p(x_{1}, x_{1\rightarrow 2}) = p(x_{2\rightarrow 1}, x_{2})$.
	\end{proposition}
	
	\subsubsection{Style-augmented Cycle Consistency} Joint distribution matching can be realized via cycle consistency constraint~\cite{zhu2017unpaired}, assuming deterministic translation models and matched marginals~\cite{li2017alice,donahue2017adversarial,dumoulin2017adversarially}. However, we note that this constraint is too strong for multimodal image translation. In fact, we prove in Appendix~\ref{app:proofs} that the translation model will degenerate to a deterministic function if cycle consistency is enforced. In the following proposition, we show that our framework admits a weaker form of cycle consistency, termed as style-augmented cycle consistency, between the image--style joint spaces, which is more \mbox{suited for multimodal image translation.}

	\begin{proposition}
		Denote $h_{1}=(x_{1}, s_{2})\in \mathcal{H}_{1}$ and $h_{2}=(x_{2}, s_{1})\in \mathcal{H}_{2}$. $h_{1}, h_{2}$ are points in the joint spaces of image and style. Our model defines a deterministic mapping $F_{1\rightarrow 2}$ from $\mathcal{H}_{1}$ to $\mathcal{H}_{2}$~(and vice versa) by $F_{1\rightarrow 2}(h_{1}) = F_{1\rightarrow 2}(x_{1}, s_{2})\triangleq(G_{2}(E^{c}_{1}(x_{1}), s_{2}), E^{s}_{1}(x_{1}))$. When optimality is achieved, we have $F_{1\rightarrow 2} = F_{2\rightarrow 1}^{-1}$.
	\end{proposition}
	
	Intuitively, style-augmented cycle consistency implies that if we translate an image to the target domain and translate it back \textit{using the original style}, we should obtain the original image. Style-augmented cycle consistency is implied by the proposed bidirectional reconstruction loss, but explicitly enforcing it could be useful for some datasets:
	\begin{align}
	\mathcal{L}^{x_{1}}_{\text{cc}}&=\mathbb{E}_{x_{1}\sim p(x_{1}), s_{2}\sim q(s_{2})}[||G_{1}(E_{2}^{c}(G_{2}(E_{1}^{c}(x_{1}), s_{2})), E_{1}^{s}(x_{1}))-x_{1}||_{1}]
	\end{align}

	\section{Experiments}
	\label{sec:experiments}	
	\subsection{Implementation Details}
	\label{sec:implementation}
\begin{figure*}[!tb]
		\centering
		\includegraphics[width=0.9\textwidth]{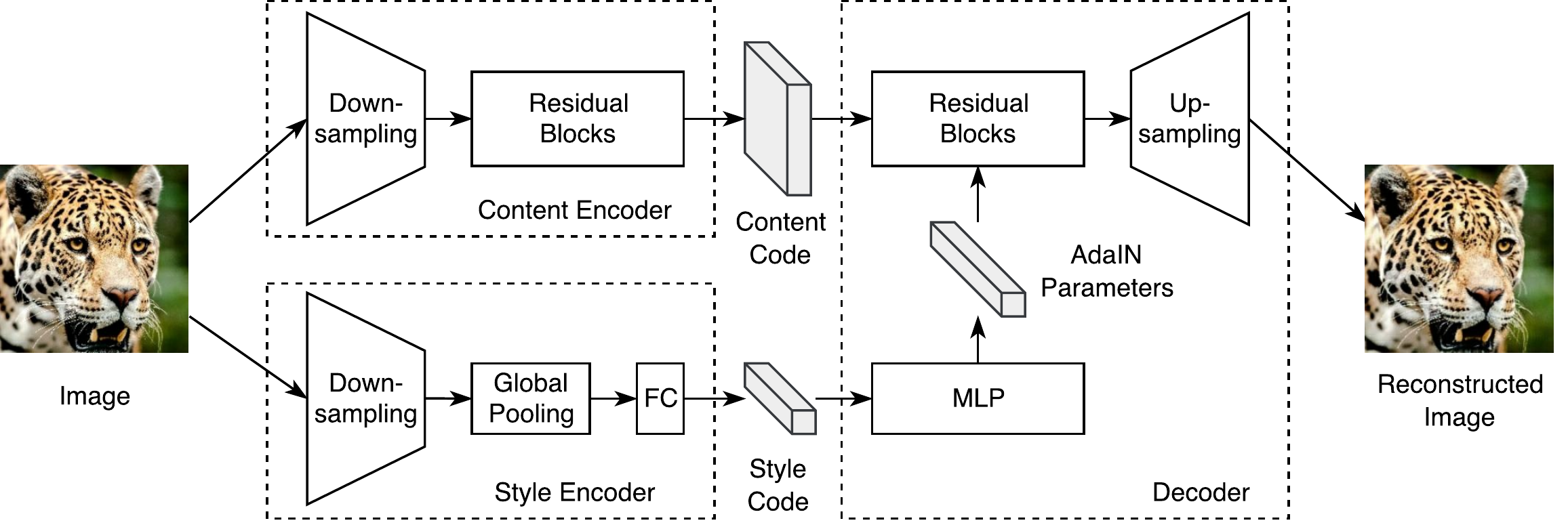}
		\caption{Our auto-encoder architecture. The content encoder consists of several strided convolutional layers followed by residual blocks. The style encoder contains several strided convolutional layers followed by a global average pooling layer and a fully connected layer. The decoder uses a MLP to produce a set of AdaIN~\cite{huang2017adain} parameters from the style code. The content code is then processed by residual blocks with AdaIN layers, and finally decoded to the image space by upsampling and convolutional layers.} 		
		\label{fig:network}
	\end{figure*}

	Fig.~\ref{fig:network} shows the architecture of our auto-encoder. It consists of a content encoder, a style encoder, and a joint decoder.
	More detailed information and hyperparameters are given in Appendix~\ref{app:hyperparameters}. We also provide an open-source implementation in PyTorch~\cite{paszke2017automatic} at \href{https://github.com/nvlabs/MUNIT}{https://github.com/nvlabs/MUNIT}.
	
	\vpara{Content encoder.} Our content encoder consists of several strided convolutional layers to downsample the input and several residual blocks~\cite{he2016deep} to further process it. All the convolutional layers are followed by \mbox{Instance Normalization~(IN)~\cite{ulyanov2017improved}.}
	
	\vpara{Style encoder.} The style encoder includes several strided convolutional layers, followed by a global average pooling layer and a fully connected~(FC) layer. We do not use IN layers in the style encoder, since IN removes the original feature mean and variance that represent important style information~\cite{huang2017adain}.

	\vpara{Decoder.} Our decoder reconstructs the input image from its content and style code. It processes the content code by a set of residual blocks and finally produces the reconstructed image by several upsampling and convolutional layers.
	Inspired by recent works that use affine transformation parameters in normalization layers to represent styles~\cite{huang2017adain,dumoulin2017learned,wang2017zm,ghiasi2017exploring}, we equip the residual blocks with Adaptive Instance Normalization~(AdaIN)~\cite{huang2017adain} layers whose parameters are dynamically generated by a multilayer perceptron~(MLP) from the style code.
	\begin{equation}
	\textrm{AdaIN}(z, \gamma, \beta)= \gamma\left(\frac{z-\mu(z)}{\sigma(z)}\right)+\beta
	\end{equation}
	where $z$ is the activation of the previous convolutional layer, $\mu$ and $\sigma$ are channel-wise mean and standard deviation, $\gamma$ and $\beta$ are parameters generated by the MLP. Note that the affine parameters are produced by a learned network, instead of computed from statistics of a pretrained network as in Huang \etal~\cite{huang2017adain}.
	
	\vpara{Discriminator.}  We use the LSGAN objective proposed by Mao \etal~\cite{mao2017least}.
	We employ multi-scale discriminators proposed by Wang~\etal~\cite{wang2018high} to guide the generators to produce both realistic details and correct global structure. 
	
	\vpara{Domain-invariant perceptual loss.} The perceptual loss, often computed as a distance in the VGG~\cite{simonyan2015very} feature space between the output and the reference image, has been shown to benefit image-to-image translation when paired supervision is available~\cite{chen2017photographic,wang2018high}. 
	In the unsupervised setting, however, we do not have a reference image in the target domain.
	We propose a modified version of perceptual loss that is more domain-invariant, so that we can use the input image as the reference. Specifically, before computing the distance, we perform Instance Normalization~\cite{ulyanov2017improved}~(without affine transformations) on the VGG features in order to remove the original feature mean and variance, which contains much domain-specific information~\cite{huang2017adain,li2016revisiting}. In Appendix~\ref{app:perceptual}, we quantitatively show that Instance Normalization can indeed make the VGG features more domain-invariant. We find the domain-invariant perceptual loss accelerates training on high-resolution~($\geq512\times512$) datasets and thus employ it on those datasets.
	

	\subsection{Evaluation Metrics}
	\label{sec:metrics}
	\vpara{Human Preference.} To compare the realism and faithfulness of translation outputs generated by different methods, we perform human perceptual study on Amazon Mechanical Turk~(AMT). Similar to Wang \etal~\cite{wang2018high}, the workers are given an input image and two translation outputs from different methods. They are then given unlimited time to select which translation output looks more accurate. For each comparison, we randomly generate $500$ questions and each question is answered by $5$ different workers. 
	
	\vpara{LPIPS Distance.} To measure translation diversity, we compute the average LPIPS distance~\cite{zhang2018unreasonable} between pairs of randomly-sampled translation outputs from the same input as in Zhu \etal~\cite{zhu2017toward}. LPIPS is given by a weighted $\mathcal{L}_{2}$ distance between deep features of images. It has been demonstrated to correlate well with human perceptual similarity~\cite{zhang2018unreasonable}. Following Zhu \etal~~\cite{zhu2017toward}, we use $100$ input images and sample $19$ output pairs per input, which amounts to $1900$ pairs in total. We use the ImageNet-pretrained \mbox{AlexNet~\cite{krizhevsky2012imagenet} as the deep feature extractor.}
	
	\vpara{(Conditional) Inception Score.} The Inception Score~(IS)~\cite{salimans2016improved} is a popular metric for image generation tasks. We propose a modified version called Conditional Inception Score~(CIS), which is more suited for evaluating multimodal image translation. When we know the number of modes in $\mathcal{X}_{2}$ as well as the ground truth mode each sample belongs to, we can train a classifier $p(y_{2}|x_{2})$ to classify an image $x_{2}$ into its mode $y_{2}$. Conditioned on a single input image $x_{1}$, the translation samples $x_{1\rightarrow 2}$ should be mode-covering~(thus $p(y_2|x_{1})= \int p(y|x_{1\rightarrow 2})p(x_{1\rightarrow 2}|x_{1}) \,dx_{1\rightarrow 2}$ should have high entropy) and each individual sample should belong to a specific mode~(thus $p(y_{2}|x_{1\rightarrow 2})$ should have low entropy). Combing these two requirements we get:
	\begin{align}
	\text{CIS} = \mathbb{E}_{x_{1}\sim p(x_{1})}[ \mathbb{E}_{x_{1\rightarrow 2}\sim p(x_{2\rightarrow 1}|x_{1})} [\text{KL}(p(y_{2}|x_{1\rightarrow 2})||p(y_{2}|x_{1}))]]
	\label{equ:cis}
	\end{align} 
	To compute the (unconditional) IS, $p(y_{2}|x_{1})$ is replaced with the unconditional class probability $p(y_{2})= \iint p(y|x_{1\rightarrow 2})p(x_{1\rightarrow 2}|x_{1})p(x_{1}) \,dx_{1}\,dx_{1\rightarrow 2}$. 
	\begin{align}
	\text{IS} = \mathbb{E}_{x_{1}\sim p(x_{1})}[ \mathbb{E}_{x_{1\rightarrow 2}\sim p(x_{2\rightarrow 1}|x_{1})} [\text{KL}(p(y_{2}|x_{1\rightarrow 2})||p(y_{2}))]]
	\label{equ:is}
	\end{align} 
	To obtain a high CIS/IS score, a model needs to generate samples that are both high-quality and diverse. While IS measures diversity of all output images, CIS measures diversity of outputs conditioned on a single input image. A model that deterministically generates a single output given an input image will receive a zero CIS score, though it might still get a high score under IS. We use the Inception-v3~\cite{szegedy2016rethinking} fine-tuned on our specific datasets as the classifier and estimate Eq.~(\ref{equ:cis}) and Eq.~(\ref{equ:is}) using $100$ input images and $100$ samples per input. 

	\subsection{Baselines}
	\vpara{UNIT~\cite{liu2017unsupervised}.} The UNIT model consists of two VAE-GANs with a fully shared latent space. The stochasticity of the translation comes from the Gaussian encoders as well as the dropout layers in the VAEs.
	
	\vpara{CycleGAN~\cite{zhu2017unpaired}.} CycleGAN consists of two residual translation networks trained with adversarial loss and cycle reconstruction loss. We use Dropout during both \mbox{training and testing to encourage diversity, as suggested in Isola \etal~\cite{isola2017image}.}
	
	\vpara{CycleGAN*~\cite{zhu2017unpaired} with noise.} To test whether we can generate multimodal outputs within the CycleGAN framework, we additionally inject noise vectors to both translation networks. We use the U-net architecture~\cite{zhu2017toward} with noise added to input, since we find the noise vectors are ignored by the residual architecture in CycleGAN~\cite{zhu2017unpaired}. Dropout is also utilized during both training and testing.
	
	
	\vpara{BicycleGAN~\cite{zhu2017toward}.} BicycleGAN is the only existing image-to-image translation model we are aware of that can generate continuous and multimodal output distributions. However, it requires paired training data. We compare our model with BicycleGAN when the dataset contains pair information. 
	
	
	
	\subsection{Datasets}
	
	\vpara{Edges $\leftrightarrow$ shoes/handbags.} We use the datasets provided by Isola~\etal~\cite{isola2017image}, Yu \etal~\cite{yu2014fine}, and Zhu~\etal\cite{zhu2016generative}, which contain images of shoes and handbags with edge maps generated by HED~\cite{xie2015holistically}. We train one model for edges $\leftrightarrow$ shoes and another for edges $\leftrightarrow$ handbags without using paired information.
	
	\vpara{Animal image translation.} We collect images from $3$ categories/domains, including house cats, big cats, and dogs. Each domain contains $4$ modes which are fine-grained categories belonging to the same parent category. Note that the modes of the images are not known during learning the translation model. 
	We learn a separate model for each pair of domains. 
	
	\vpara{Street scene images.} We experiment with two street scene translation tasks: 
	\begin{itemize}
		\item \textbf{Synthetic $\leftrightarrow$ real.} We perform translation between synthetic images in the SYNTHIA dataset~\cite{ros2016synthia} and real-world images in the Cityscape dataset~\cite{cordts2016cityscapes}. For the SYNTHIA dataset, we use the SYNTHIA-Seqs subset which contains images in different seasons, weather, and illumination conditions.
		\item \textbf{Summer $\leftrightarrow$ winter.} We use the dataset from Liu et al.~\cite{liu2017unsupervised}, which contains summer and winter street images extracted from real-world driving videos.
	\end{itemize}
	\vpara{Yosemite summer $\leftrightarrow$ winter (HD).} We collect a new high-resolution dataset containing $3253$ summer photos and $2385$ winter photos of Yosemite. The images are downsampled such that the shortest side of each image is $1024$ pixels.

	\begin{figure*}[!tb]
		\centering
		\small
		\newcommand{\sizea}{0.1039\linewidth}
		\lstackon[47pt]{\lstackon[39pt]{\includegraphics[width=\sizea]{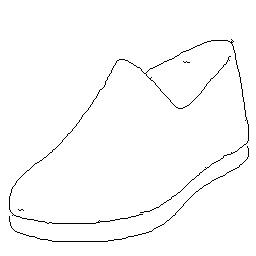}}{\tiny \& GT}}{{\tiny Input}}
		\lstackon[47pt]{\lstackon[39pt]{\includegraphics[width=\sizea]{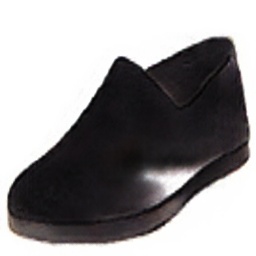}}{}}{\tiny UNIT}
		\lstackon[47pt]{\lstackon[39pt]{\includegraphics[width=\sizea]{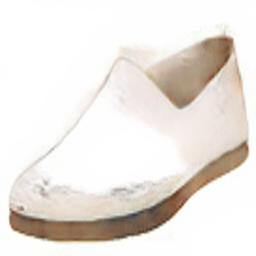}}{}}{\tiny CycleGAN}
		\lstackon[47pt]{\lstackon[39pt]{\includegraphics[width=\sizea]{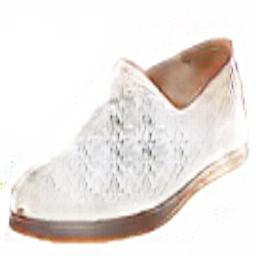}}{\tiny with noise}}{{\tiny CycleGAN*}}
		\lstackon[47pt]{\lstackon[39pt]{\includegraphics[width=\sizea]{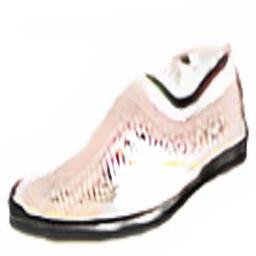}}{\tiny w/o $\mathcal{L}^{x}_{\text{recon}}$}}{\tiny MUNIT}
		\lstackon[47pt]{\lstackon[39pt]{\includegraphics[width=\sizea]{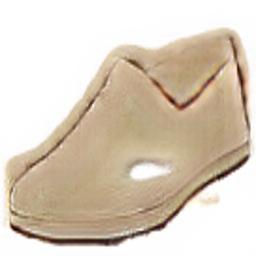}}{\tiny w/o $\mathcal{L}^{c}_{\text{recon}}$}}{\tiny MUNIT}
		\lstackon[47pt]{\lstackon[39pt]{\includegraphics[width=\sizea]{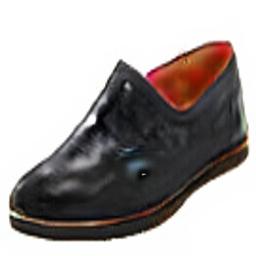}}{\tiny w/o $\mathcal{L}^{s}_{\text{recon}}$}}{\tiny MUNIT}
		\lstackon[47pt]{\lstackon[39pt]{\includegraphics[width=\sizea]{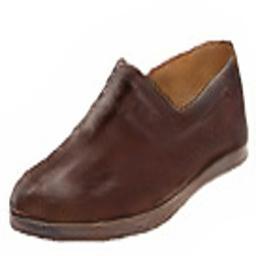}}{\tiny (ours)}}{\tiny MUNIT}
		\lstackon[47pt]{\lstackon[39pt]{\includegraphics[width=\sizea]{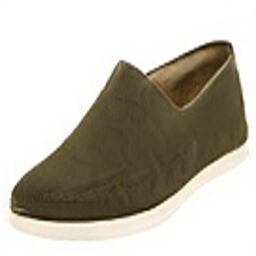}}{\tiny GAN}}{\tiny Bicycle-}\\
		\includegraphics[width=\sizea]{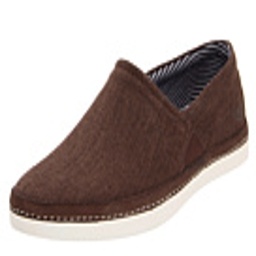}
		\includegraphics[width=\sizea]{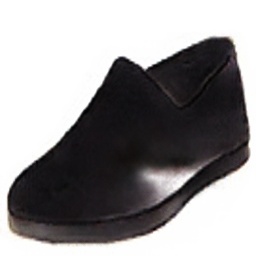}
		\includegraphics[width=\sizea]{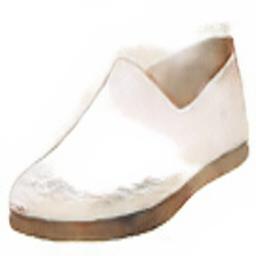}
		\includegraphics[width=\sizea]{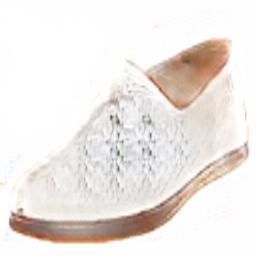}
		\includegraphics[width=\sizea]{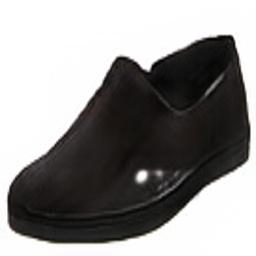}
		\includegraphics[width=\sizea]{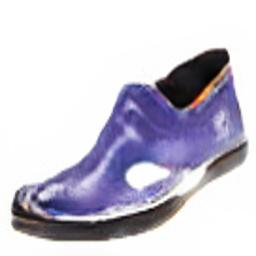}
		\includegraphics[width=\sizea]{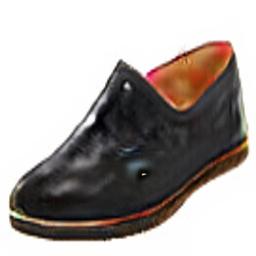}
		\includegraphics[width=\sizea]{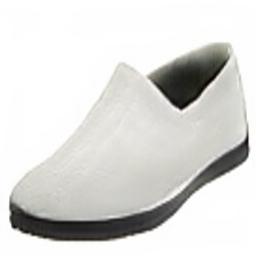}
		\includegraphics[width=\sizea]{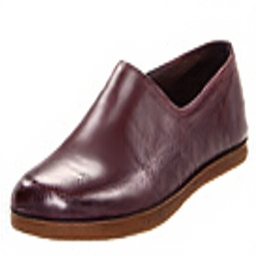}
		\phantom{\includegraphics[width=\sizea]{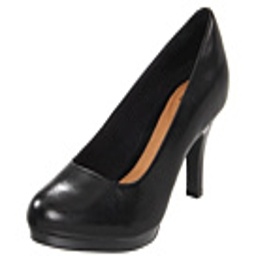}}
		\includegraphics[width=\sizea]{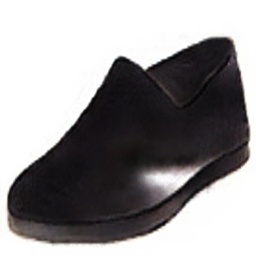}
		\includegraphics[width=\sizea]{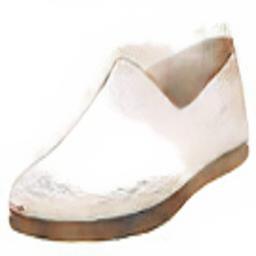}
		\includegraphics[width=\sizea]{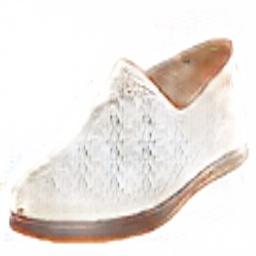}
		\includegraphics[width=\sizea]{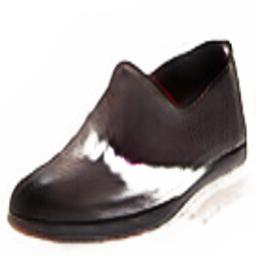}
		\includegraphics[width=\sizea]{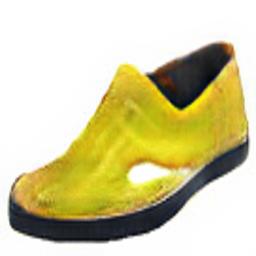}
		\includegraphics[width=\sizea]{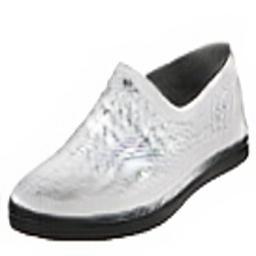}
		\includegraphics[width=\sizea]{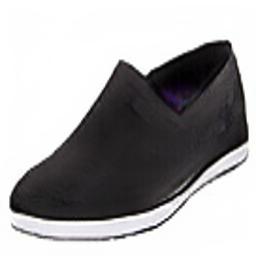}
		\includegraphics[width=\sizea]{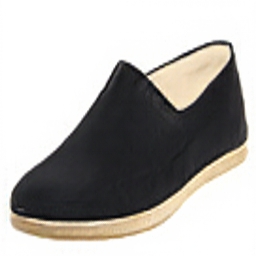}
		\caption{Qualitative comparison on edges $\rightarrow$ shoes. The first column shows the input and ground truth output. Each following column shows $3$ random outputs from a method.}
		\label{fig:comparison}
	\end{figure*}

	\begin{table}[!tb]
		\addtolength{\tabcolsep}{8pt}
		\renewcommand\arraystretch{1.3}
		\centering
		\small
		\begin{threeparttable}
			\caption{Quantitative evaluation on edges $\rightarrow$ shoes/handbags. The diversity score is the average LPIPS distance~\cite{zhang2018unreasonable}. The quality score is the human preference score, the percentage a method is preferred over MUNIT. For both metrics, the higher the better.\label{tab:edges}}
			\begin{tabular}{|c|c|c|c|c|c|c|c|}
				\hline
				& \multicolumn{2}{c|}{edges $\rightarrow$ shoes} & \multicolumn{2}{c|}{edges $\rightarrow$ handbags} \\  \cline{2-5}
				& \ Quality\ \  & Diversity &  \ Quality\ \  & Diversity \\ 
				\hline
				UNIT~\cite{liu2017unsupervised} & 37.4\% & 0.011 & 37.3\%  & 0.023 \\
				CycleGAN~\cite{zhu2017unpaired} & 36.0\% & 0.010 & 40.8\% & 0.012 \\
				CycleGAN*~\cite{zhu2017unpaired} with noise & 29.5\% & 0.016 & 45.1\% & 0.011 \\
				\hline
				MUNIT w/o $\mathcal{L}^{x}_{\text{recon}}$ & 6.0\% & 0.213 & 29.0\% & 0.191 \\
				MUNIT w/o $\mathcal{L}^{c}_{\text{recon}}$ & 20.7\% & 0.172 & 9.3\% & 0.185\\
				MUNIT w/o $\mathcal{L}^{s}_{\text{recon}}$ & 28.6\% & 0.070 & 24.6\% & 0.139 \\
				MUNIT  & 50.0\% & 0.109 & 50.0\% & 0.175 \\
				\hline
				BicycleGAN~\cite{zhu2017toward}$^\dag$ & 56.7\%  & 0.104 & 51.2\% & 0.140 \\
				\hline
				Real data & N/A & 0.293 & N/A & 0.371 \\
				\hline
			\end{tabular}
			\begin{tablenotes}
				\item [\dag] Trained with paired supervision.
			\end{tablenotes}
		\end{threeparttable}
	\end{table}
	
	\subsection{Results}
	First, we qualitatively compare MUNIT with the four baselines above, and three variants of MUNIT that ablate $\mathcal{L}^{x}_{\text{recon}}$, $\mathcal{L}^{c}_{\text{recon}}$, $\mathcal{L}^{s}_{\text{recon}}$ respectively. Fig.~\ref{fig:comparison} shows example results on edges $\rightarrow$ shoes. Both UNIT and CycleGAN (with or without noise) fail to generate diverse outputs, despite the injected randomness. Without $\mathcal{L}^{x}_{\text{recon}}$ or $\mathcal{L}^{c}_{\text{recon}}$, the image quality of MUNIT  is unsatisfactory. Without $\mathcal{L}^{s}_{\text{recon}}$, the model suffers from partial mode collapse, with many outputs being almost identical~(\textit{e.g.}, the first two rows). Our full model produces images that are both diverse and realistic, similar to BicycleGAN but does not need supervision. 
	
\begin{figure*}[!tb]
		\centering
		\small
		\newcommand{\sizea}{0.091\linewidth}
		
		\setlength{\arrayrulewidth}{.5pt}%
		\setlength{\tabcolsep}{0pt}
		\renewcommand{\arraystretch}{0}
		\lstackunder[70pt]{\begin{tabular}{cc;{2pt/2pt}c}
				\lstackon[63pt]{\begin{minipage}[h]{\sizea}
						\includegraphics[width=1\linewidth]{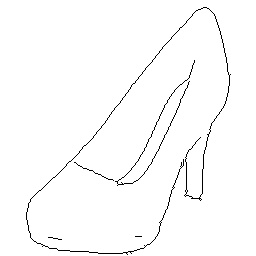}\vspace{0.2cm}
						\includegraphics[width=1\linewidth]{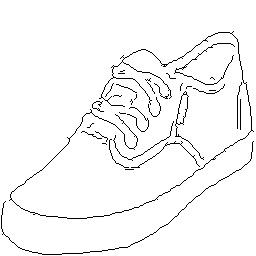}\vspace{0.2cm}
						\includegraphics[width=1\linewidth]{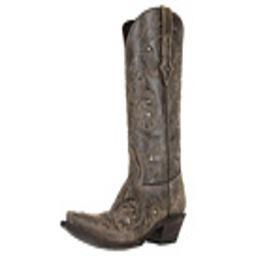}
					\end{minipage}
				}{Input} &
				\lstackon[63pt]{\begin{minipage}[h]{\sizea}
						\includegraphics[width=1\linewidth]{figures/edges2shoes_gt0.jpg}\vspace{0.2cm}
						\includegraphics[width=1\linewidth]{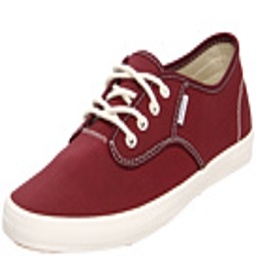}\vspace{0.2cm}
						\includegraphics[width=1\linewidth]{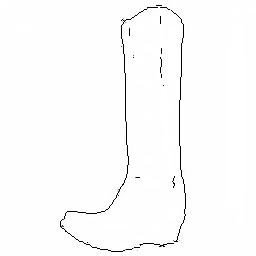}
				\end{minipage}}{GT}\ \ &
				\ \lstackon[63pt]{\begin{minipage}[h]{\sizea}
						\includegraphics[width=1\linewidth]{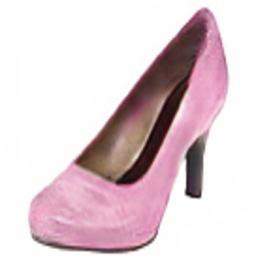}\vspace{0.2cm}
						\includegraphics[width=1\linewidth]{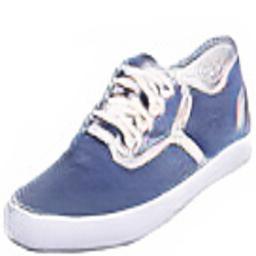}\vspace{0.2cm}
						\includegraphics[width=1\linewidth]{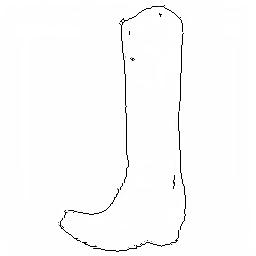}
					\end{minipage}\begin{minipage}[h]{\sizea}
						\includegraphics[width=1\linewidth]{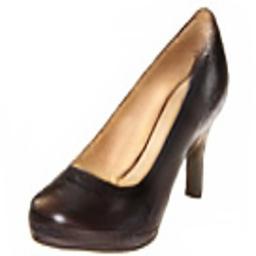}\vspace{0.2cm}
						\includegraphics[width=1\linewidth]{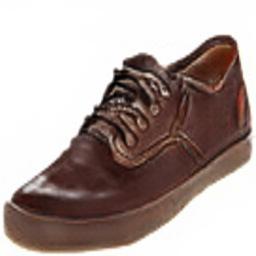}\vspace{0.2cm}
						\includegraphics[width=1\linewidth]{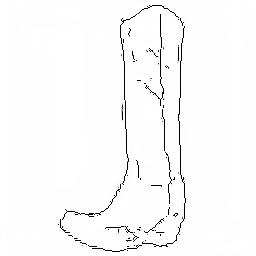}
					\end{minipage}\begin{minipage}[h]{\sizea}
						\includegraphics[width=1\linewidth]{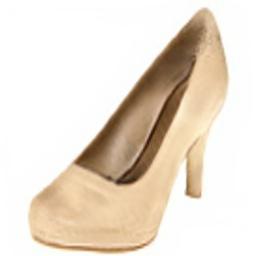}\vspace{0.2cm}
						\includegraphics[width=1\linewidth]{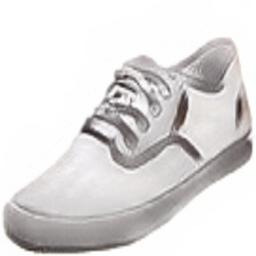}\vspace{0.2cm}
						\includegraphics[width=1\linewidth]{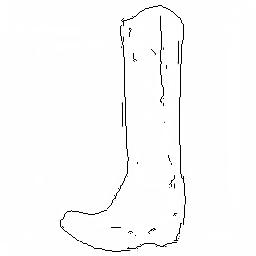}
				\end{minipage}}{Sample translations}
		\end{tabular}}{(a) edges $\leftrightarrow$ shoes}\hspace{0.3cm}
		\lstackunder[70pt]{\begin{tabular}{cc;{2pt/2pt}c}
				\lstackon[63pt]{\begin{minipage}[h]{\sizea}
						\includegraphics[width=1\linewidth]{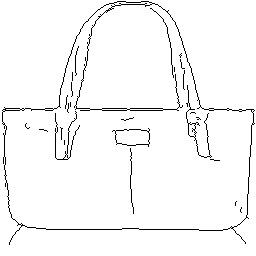}\vspace{0.2cm}
						\includegraphics[width=1\linewidth]{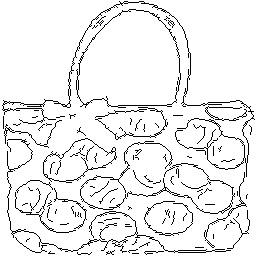}\vspace{0.2cm}
						\includegraphics[width=1\linewidth]{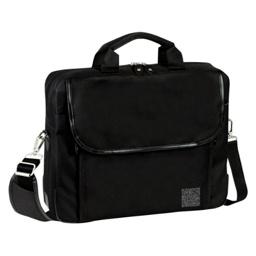}
					\end{minipage}
				}{Input} &
				\lstackon[63pt]{\begin{minipage}[h]{\sizea}
						\includegraphics[width=1\linewidth]{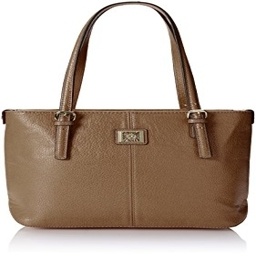}\vspace{0.2cm}
						\includegraphics[width=1\linewidth]{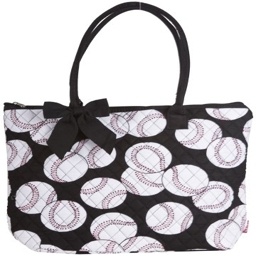}\vspace{0.2cm}
						\includegraphics[width=1\linewidth]{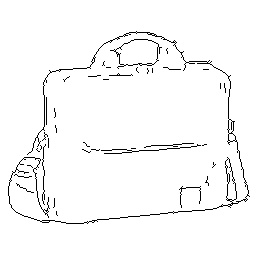}
				\end{minipage}}{GT}\ \ &
				\ \lstackon[63pt]{\begin{minipage}[h]{\sizea}
						\includegraphics[width=1\linewidth]{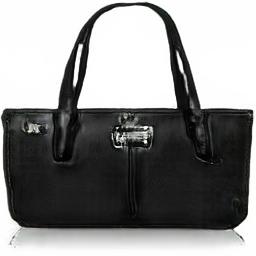}\vspace{0.2cm}
						\includegraphics[width=1\linewidth]{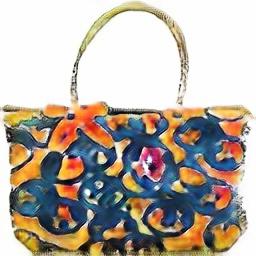}\vspace{0.2cm}
						\includegraphics[width=1\linewidth]{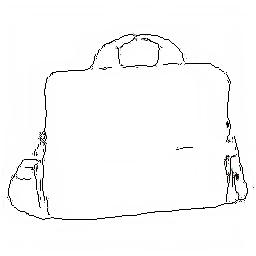}
					\end{minipage}\begin{minipage}[h]{\sizea}
						\includegraphics[width=1\linewidth]{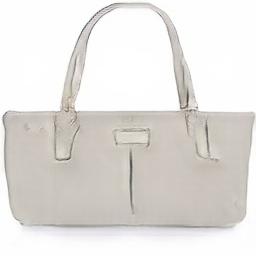}\vspace{0.2cm}
						\includegraphics[width=1\linewidth]{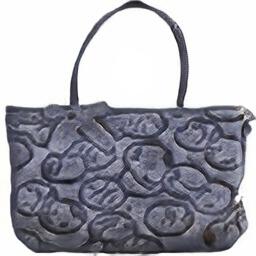}\vspace{0.2cm}
						\includegraphics[width=1\linewidth]{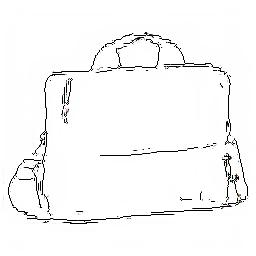}
					\end{minipage}\begin{minipage}[h]{\sizea}
						\includegraphics[width=1\linewidth]{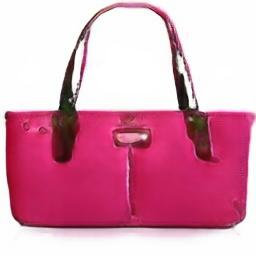}\vspace{0.2cm}
						\includegraphics[width=1\linewidth]{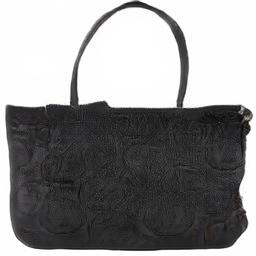}\vspace{0.2cm}
						\includegraphics[width=1\linewidth]{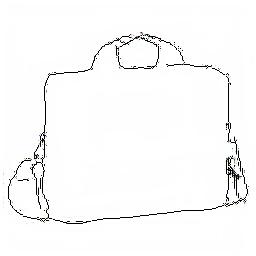}
				\end{minipage}}{Sample translations}
		\end{tabular}}{(b) edges $\leftrightarrow$ handbags}	
		\vspace{-0.1in}
		\caption{Example results of (a) edges $\leftrightarrow$ shoes and (b) edges $\leftrightarrow$ handbags.}
		\label{fig:examples_edges}
	\end{figure*}

	The qualitative observations above are confirmed by quantitative evaluations. We use human preference to measure quality and LPIPS distance to evaluate diversity, as described in Sec.~\ref{sec:metrics}. We conduct this experiment on the task of edges $\rightarrow$ shoes/handbags. As shown in Table~\ref{tab:edges}, UNIT and CycleGAN produce very little diversity according to LPIPS distance. Removing $\mathcal{L}^{x}_{\text{recon}}$ or $\mathcal{L}^{c}_{\text{recon}}$ from MUNIT leads to significantly worse quality. Without $\mathcal{L}^{s}_{\text{recon}}$, both quality and diversity deteriorate. The full model obtains quality and diversity comparable to the fully supervised BicycleGAN, and significantly better than all unsupervised baselines.
	In Fig.~\ref{fig:examples_edges}, we show more example results on edges $\leftrightarrow$ shoes/handbags. 
	
	\begin{figure*}[!tb]
		\centering
		\small
		\newcommand{\sizea}{0.1105\linewidth}
		
		\setlength{\arrayrulewidth}{.5pt}%
		\setlength{\tabcolsep}{3pt}
		\renewcommand{\arraystretch}{0}
		\lstackunder[33pt]{\begin{tabular}{c;{2pt/2pt}c}
				\lstackon[43pt]{\includegraphics[width=\sizea]{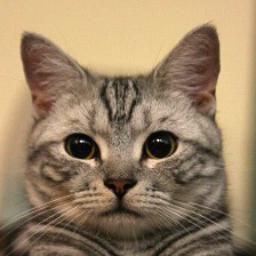}}{Input} & \lstackon[43pt]{\includegraphics[width=\sizea]{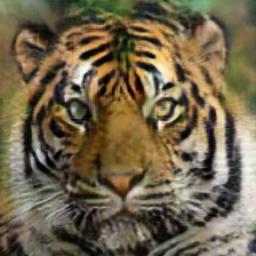}\ \includegraphics[width=\sizea]{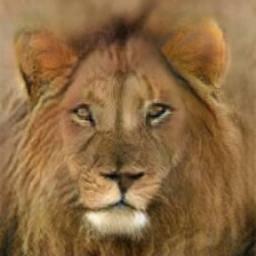}\ \includegraphics[width=\sizea]{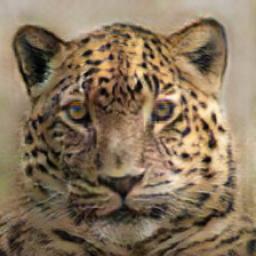}}{Sample translations}
		\end{tabular}}{(a) house cats $\rightarrow$ big cats}\vspace{0.1cm}
		\lstackunder[33pt]{\begin{tabular}{c;{2pt/2pt}c}
				\lstackon[43pt]{\includegraphics[width=\sizea]{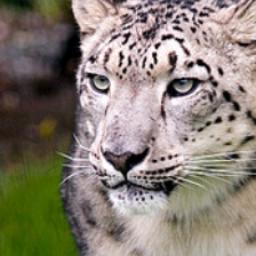}}{Input} & \lstackon[43pt]{\includegraphics[width=\sizea]{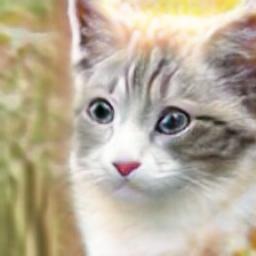}\ \includegraphics[width=\sizea]{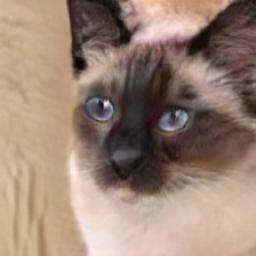}\ \includegraphics[width=\sizea]{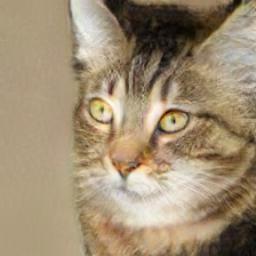}}{Sample translations}
		\end{tabular}}{(b) big cats $\rightarrow$ house cats}
		\lstackunder[27pt]{\begin{tabular}{c;{2pt/2pt}c}
				\includegraphics[width=\sizea]{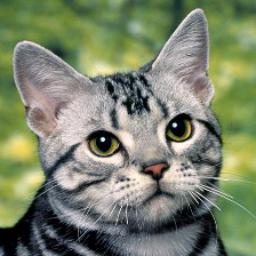} & {\includegraphics[width=\sizea]{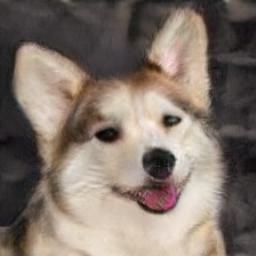}\ \includegraphics[width=\sizea]{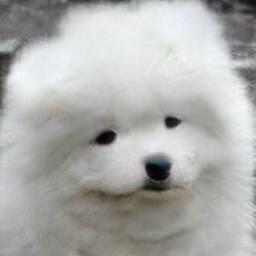}\ \includegraphics[width=\sizea]{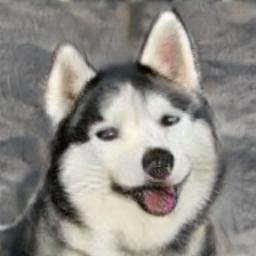}}
		\end{tabular}}{(c) house cats $\rightarrow$ dogs}\vspace{0.1cm}
		\lstackunder[27pt]{\begin{tabular}{c;{2pt/2pt}c}
				\includegraphics[width=\sizea]{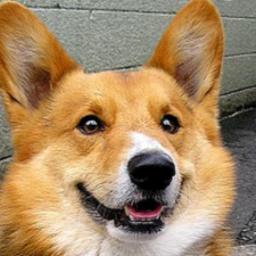} & {\includegraphics[width=\sizea]{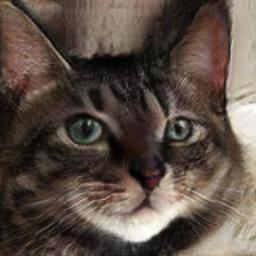}\ \includegraphics[width=\sizea]{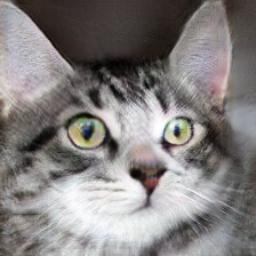}\ \includegraphics[width=\sizea]{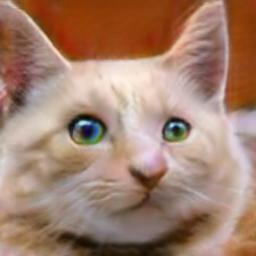}}
		\end{tabular}}{(d) dogs $\rightarrow$ house cats}
		
		\lstackunder[27pt]{\begin{tabular}{c;{2pt/2pt}c}
				\includegraphics[width=\sizea]{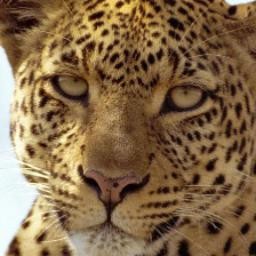} & {\includegraphics[width=\sizea]{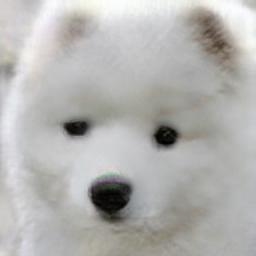}\ \includegraphics[width=\sizea]{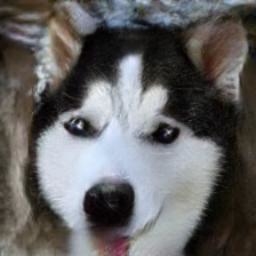}\ \includegraphics[width=\sizea]{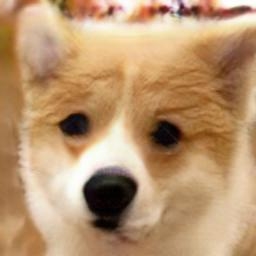}}
		\end{tabular}}{(e) big cats $\rightarrow$ dogs}
		\lstackunder[27pt]{\begin{tabular}{c;{2pt/2pt}c}
				\includegraphics[width=\sizea]{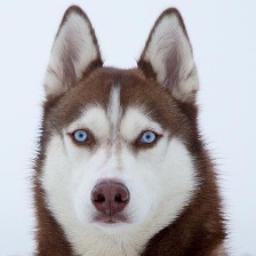} & {\includegraphics[width=\sizea]{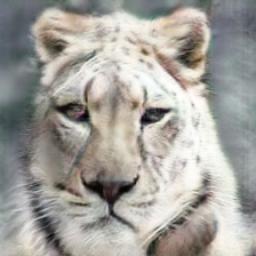}\ \includegraphics[width=\sizea]{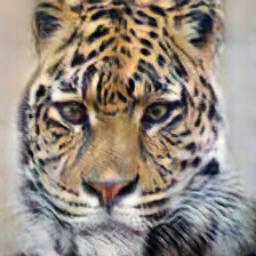}\ \includegraphics[width=\sizea]{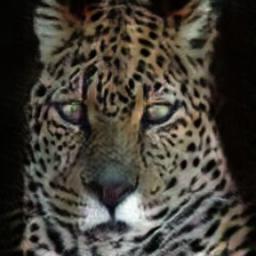}}
		\end{tabular}}{(f) dogs $\rightarrow$ big cats}
		\caption{Example results of animal image translation.}
		\label{fig:examples_animals}
	\end{figure*}
	
		\begin{figure*}[!tb]
		\centering
		\small
		\newcommand{\sizea}{0.238\linewidth}
		
		\setlength{\arrayrulewidth}{.5pt}%
		\setlength{\tabcolsep}{3pt}
		\renewcommand{\arraystretch}{0}
		\lstackunder[42pt]{\begin{tabular}{c;{2pt/2pt}c}
				\lstackon[60pt]{\includegraphics[width=\sizea]{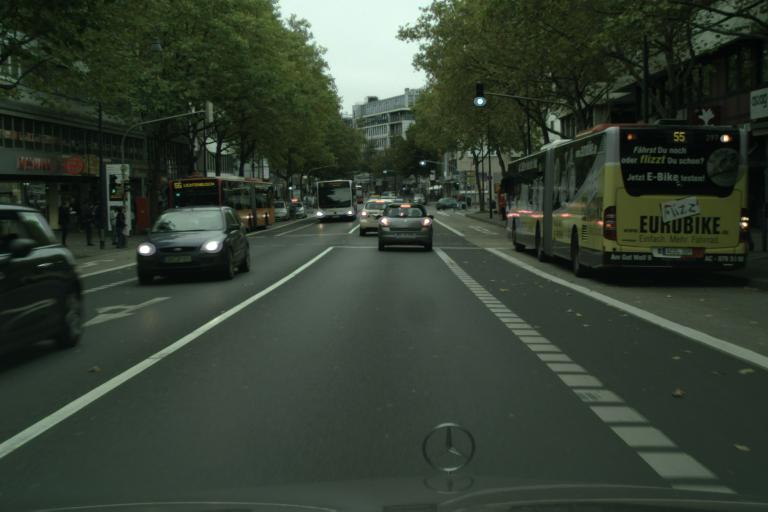}}{Input} & \lstackon[60pt]{\includegraphics[width=\sizea]{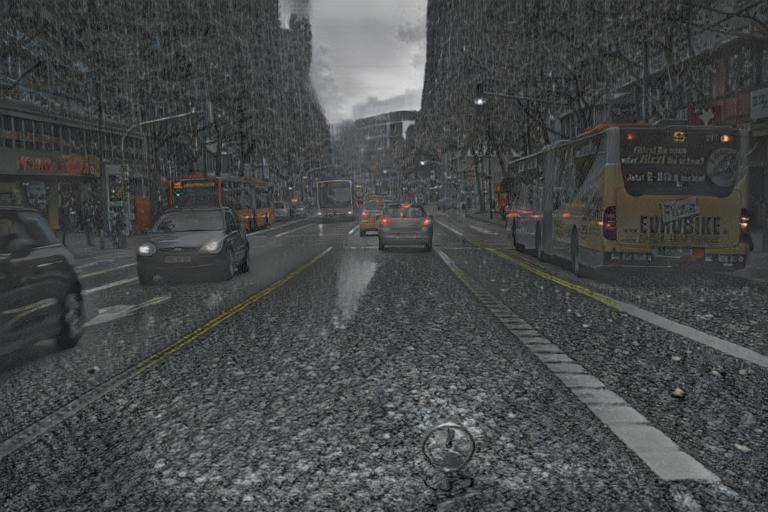}\ \includegraphics[width=\sizea]{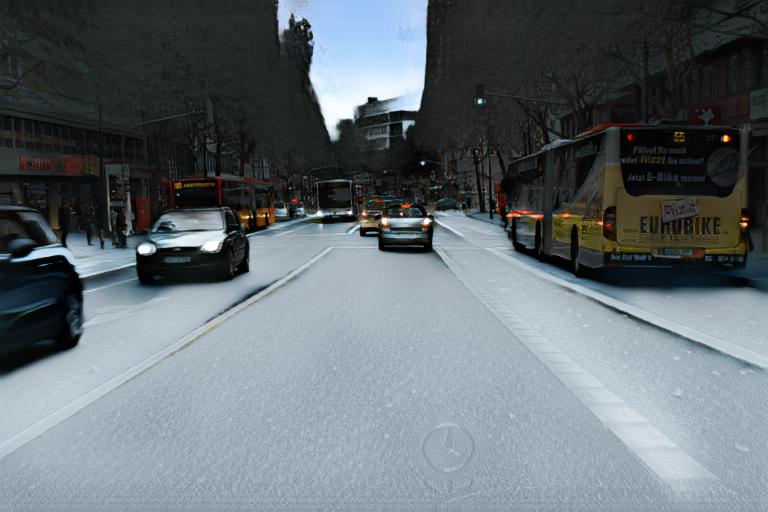}\ \includegraphics[width=\sizea]{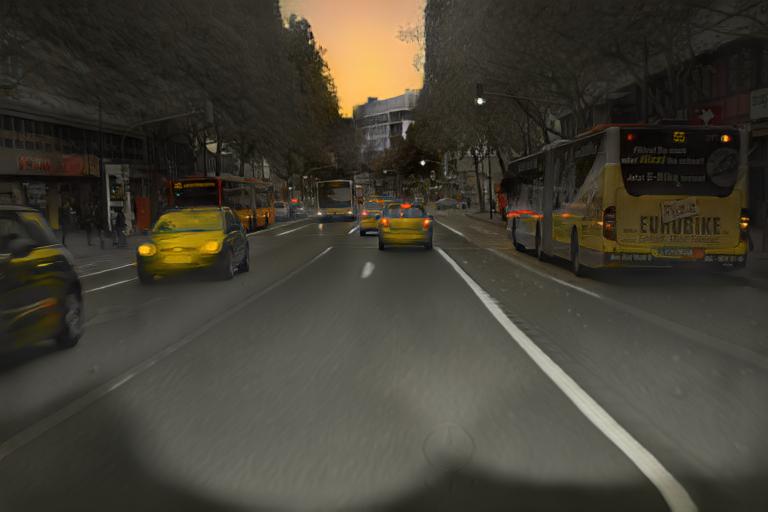}}{Sample translations}
		\end{tabular}}{(a) Cityscape $\rightarrow$ SYNTHIA}\vspace{0.1cm}
		\lstackunder[36pt]{\begin{tabular}{c;{2pt/2pt}c}
				\includegraphics[width=\sizea]{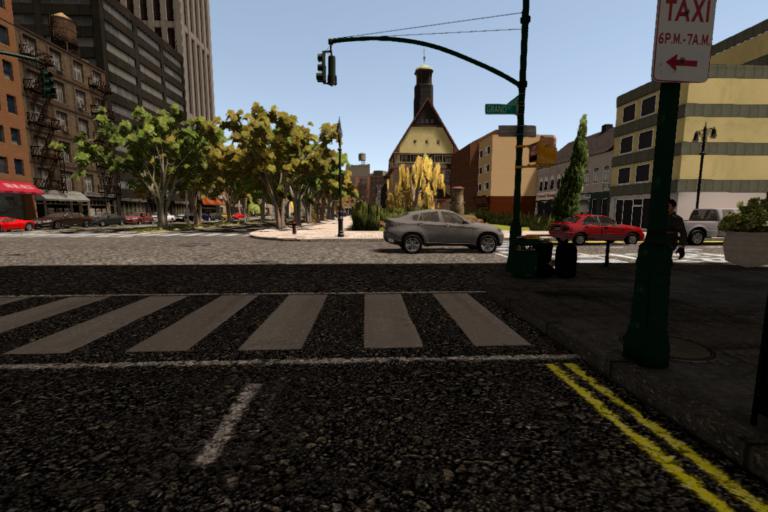} & {\includegraphics[width=\sizea]{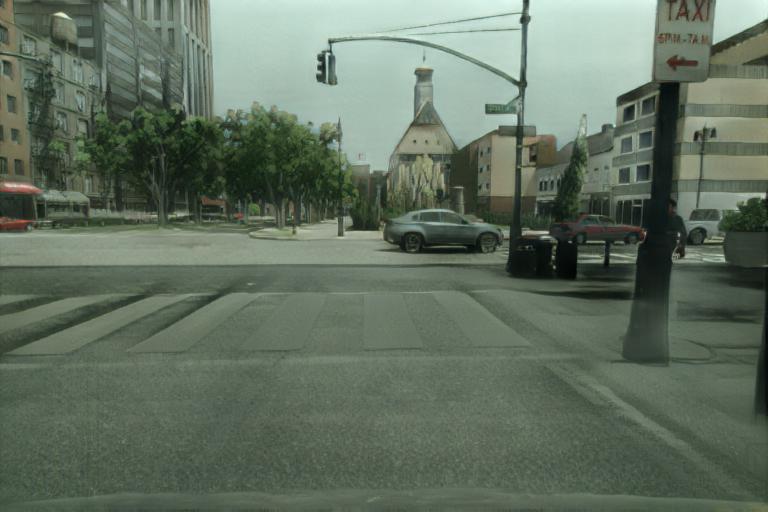}\ \includegraphics[width=\sizea]{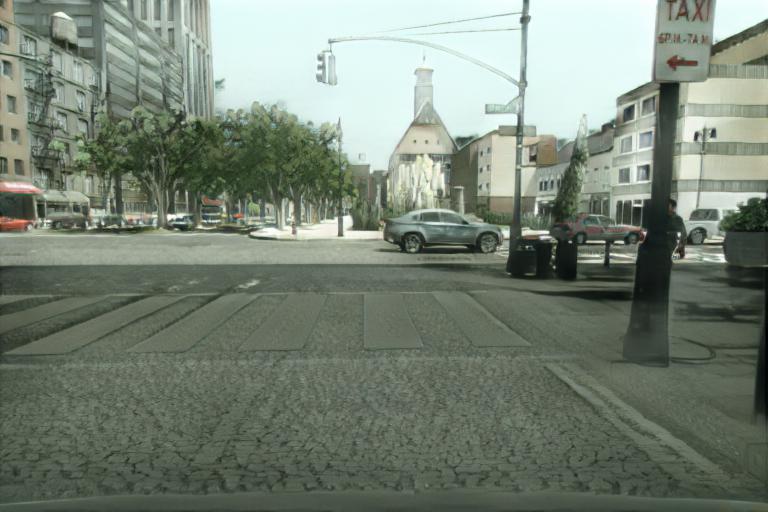}\ \includegraphics[width=\sizea]{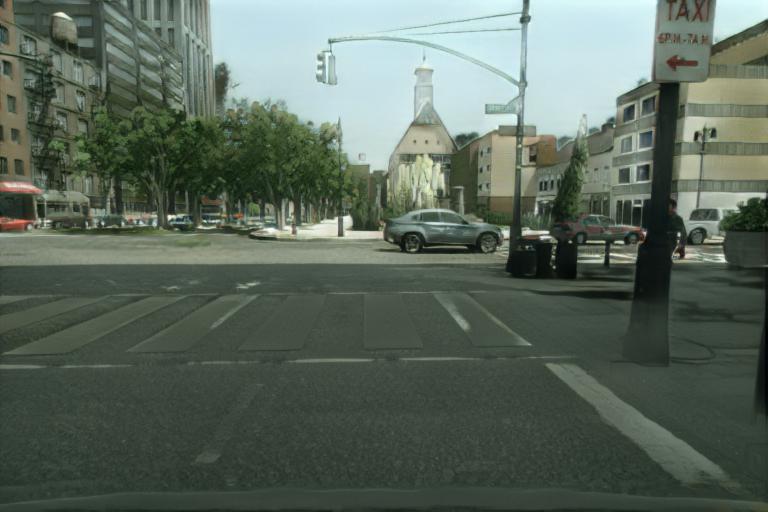}}
		\end{tabular}}{(b) SYNTHIA $\rightarrow$ Cityscape}	
		\lstackunder[36pt]{\begin{tabular}{c;{2pt/2pt}c}
				\includegraphics[width=\sizea]{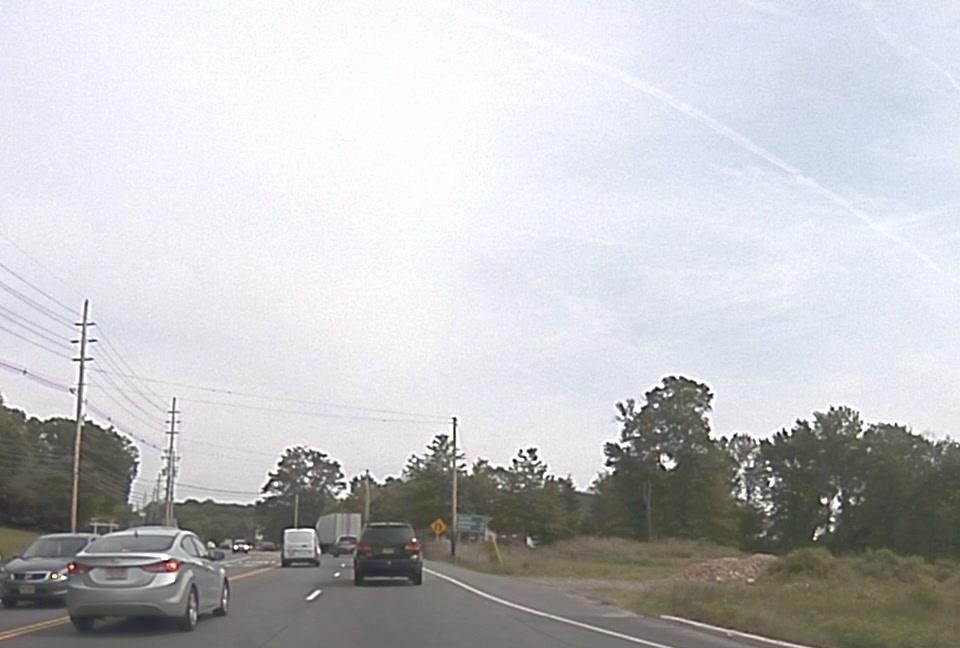} & {\includegraphics[width=\sizea]{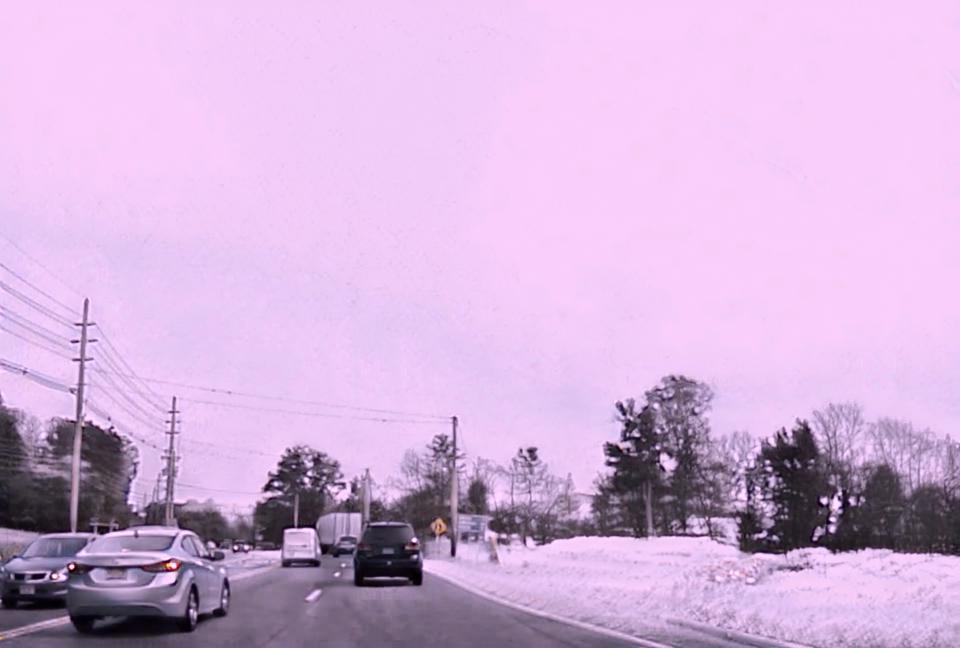}\ \includegraphics[width=\sizea]{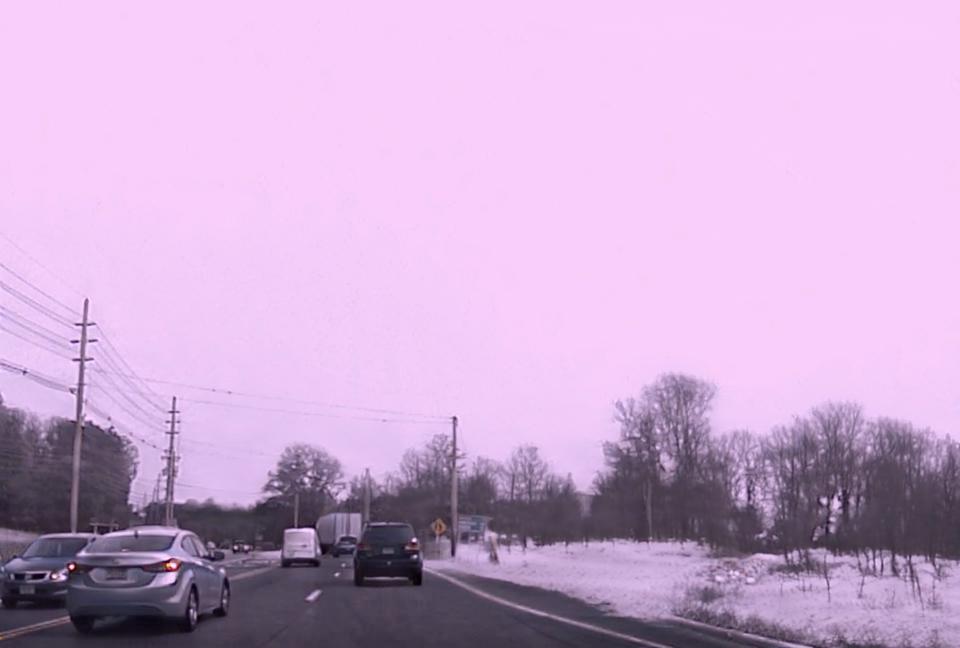}\ \includegraphics[width=\sizea]{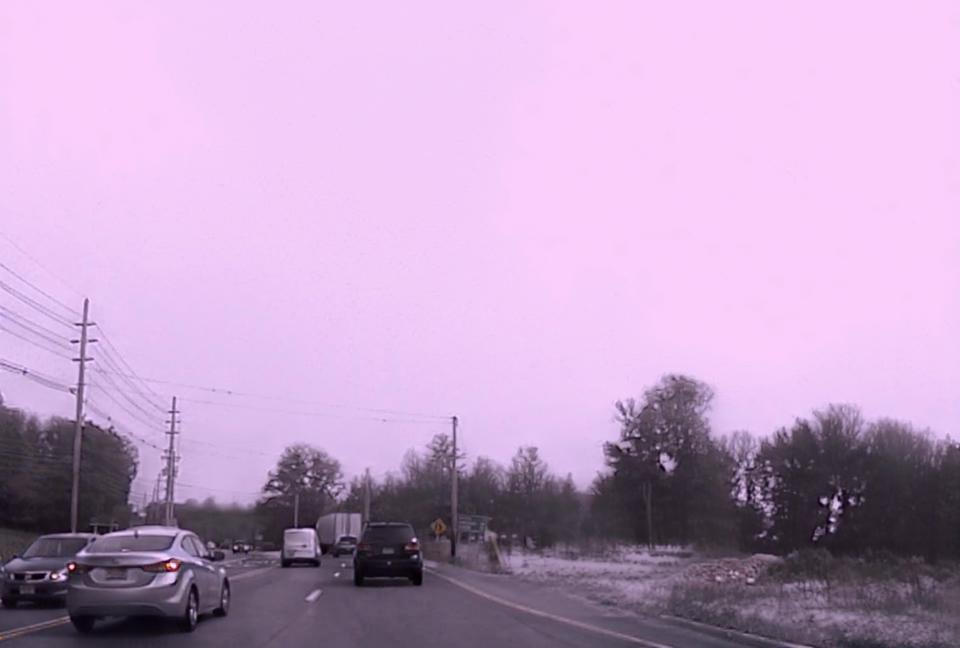}}
		\end{tabular}}{(c) summer $\rightarrow$ winter}	
		\lstackunder[36pt]{\begin{tabular}{c;{2pt/2pt}c}
				\includegraphics[width=\sizea]{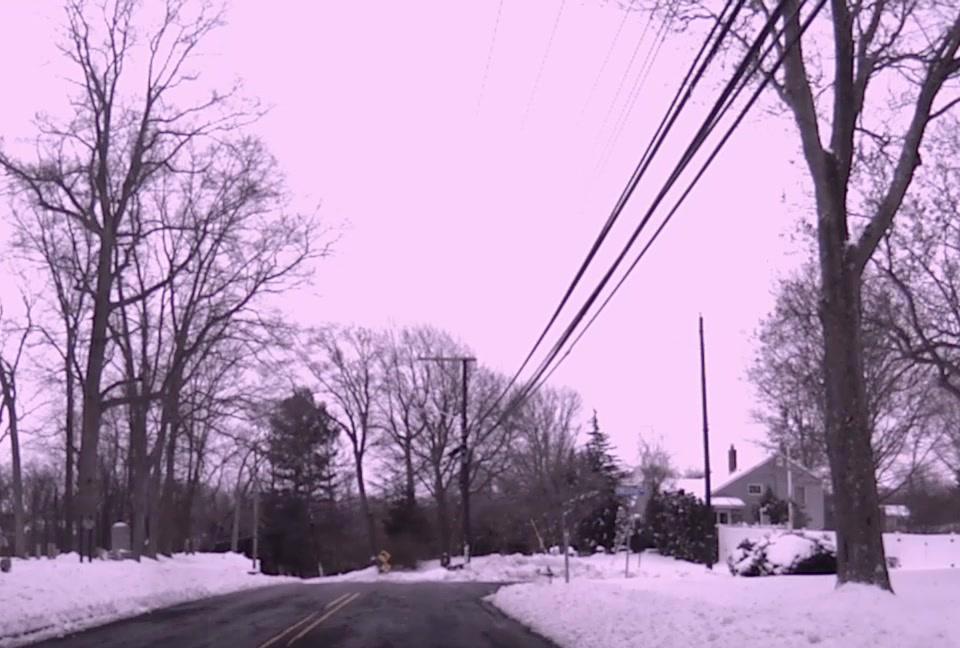} & {\includegraphics[width=\sizea]{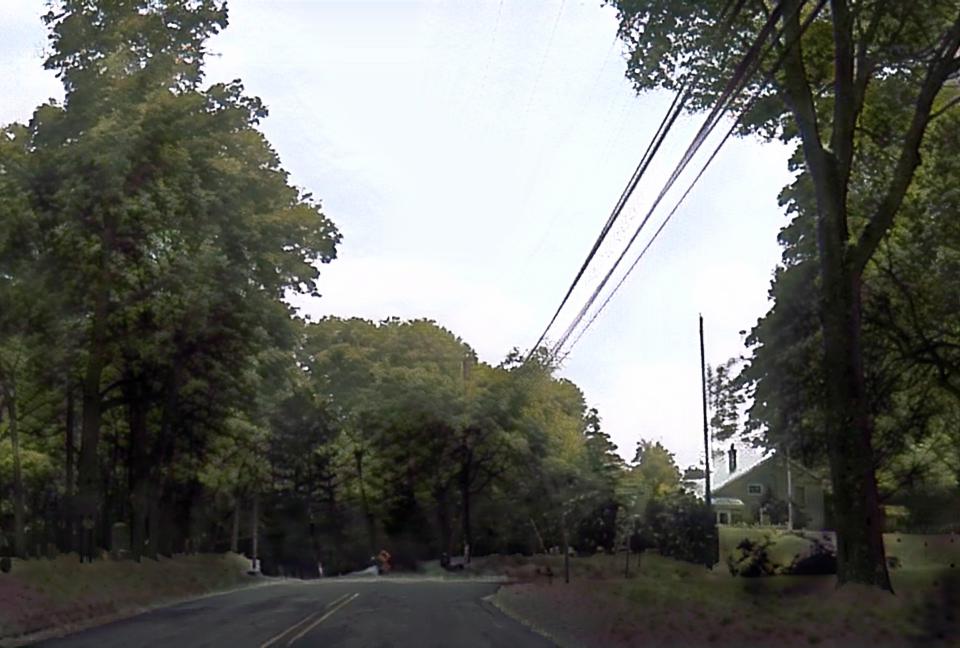}\ \includegraphics[width=\sizea]{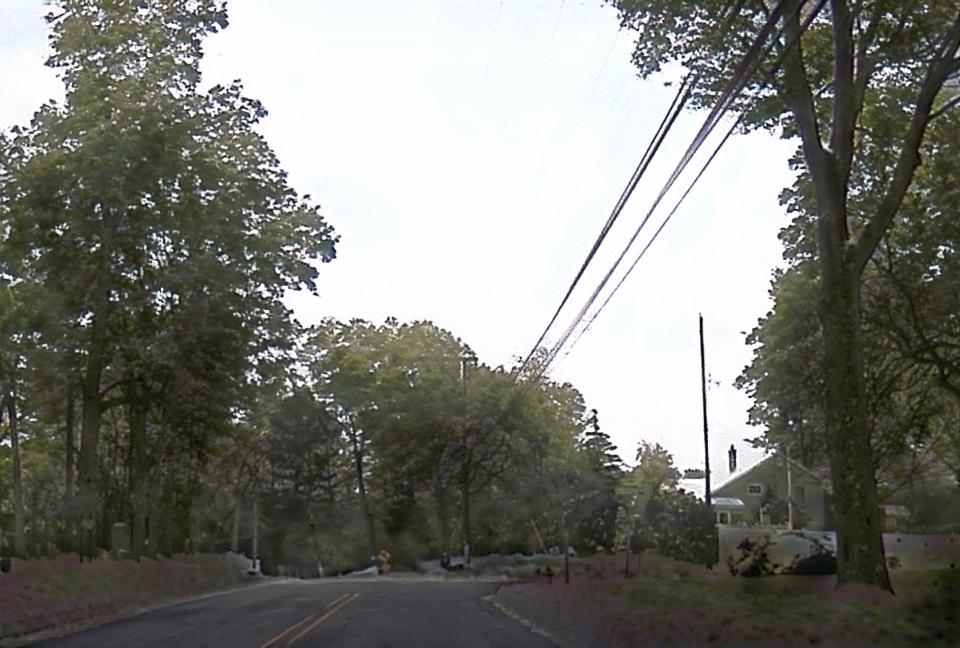}\ \includegraphics[width=\sizea]{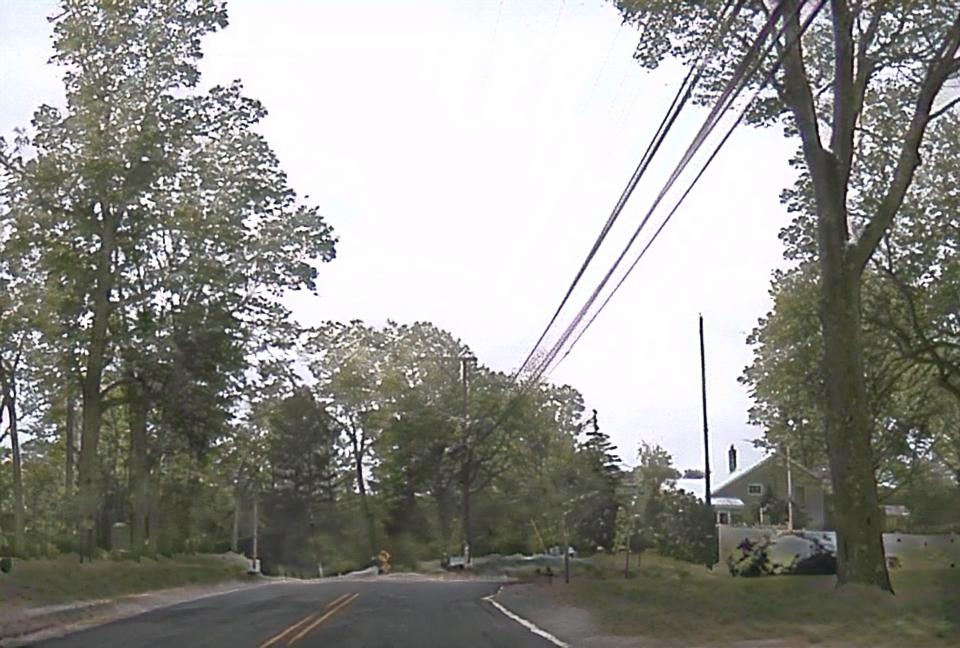}}
		\end{tabular}}{(d) winter $\rightarrow$ summer}	
		\vspace{-0.1in}
		\caption{Example results on street scene translations.}
		\vspace{-0.2in}
		\label{fig:examples_street}
	\end{figure*}

	\begin{figure*}[!tb]
		\centering
		\small
		\newcommand{\sizea}{0.238\linewidth}
		
		\setlength{\arrayrulewidth}{.5pt}%
		\setlength{\tabcolsep}{3pt}
		\renewcommand{\arraystretch}{0}
		\lstackunder[40pt]{\begin{tabular}{c;{2pt/2pt}c}
				\lstackon[53pt]{\includegraphics[width=\sizea]{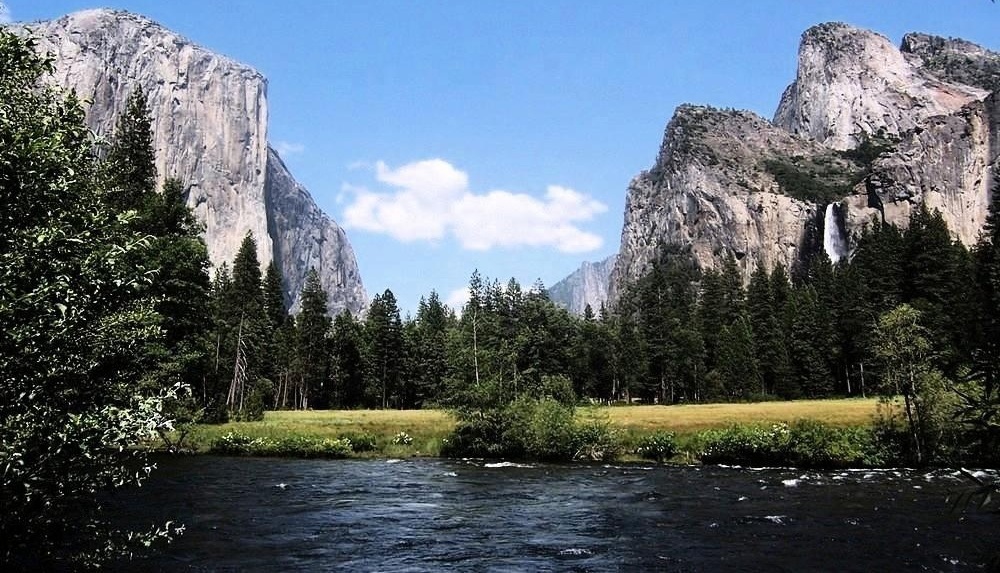}}{Input} & \lstackon[53pt]{\includegraphics[width=\sizea]{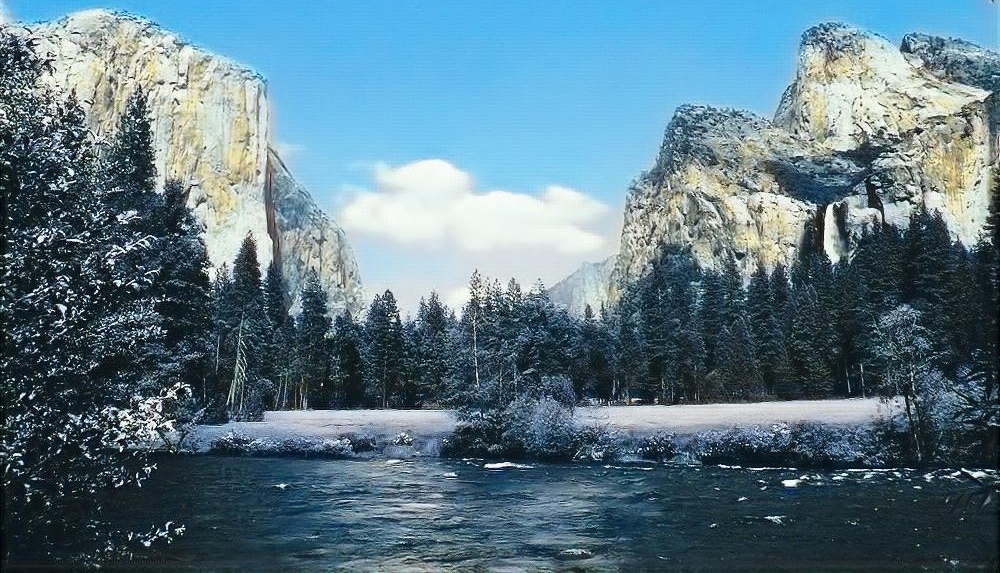}\ \includegraphics[width=\sizea]{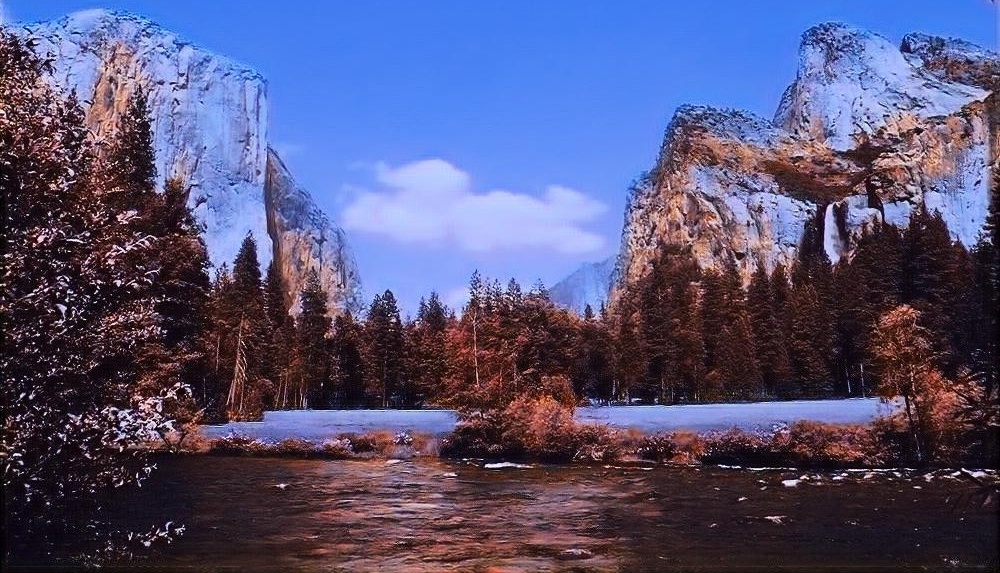}\ \includegraphics[width=\sizea]{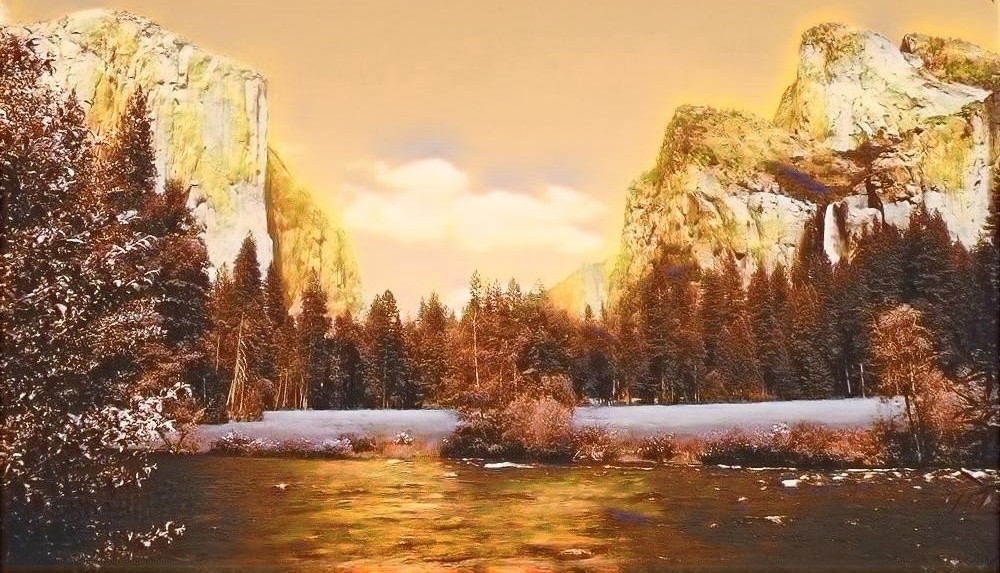}}{Sample translations}
		\end{tabular}}{(a) Yosemite summer $\rightarrow$ winter}\vspace{0.1cm}
		\lstackunder[33pt]{\begin{tabular}{c;{2pt/2pt}c}
				\includegraphics[width=\sizea]{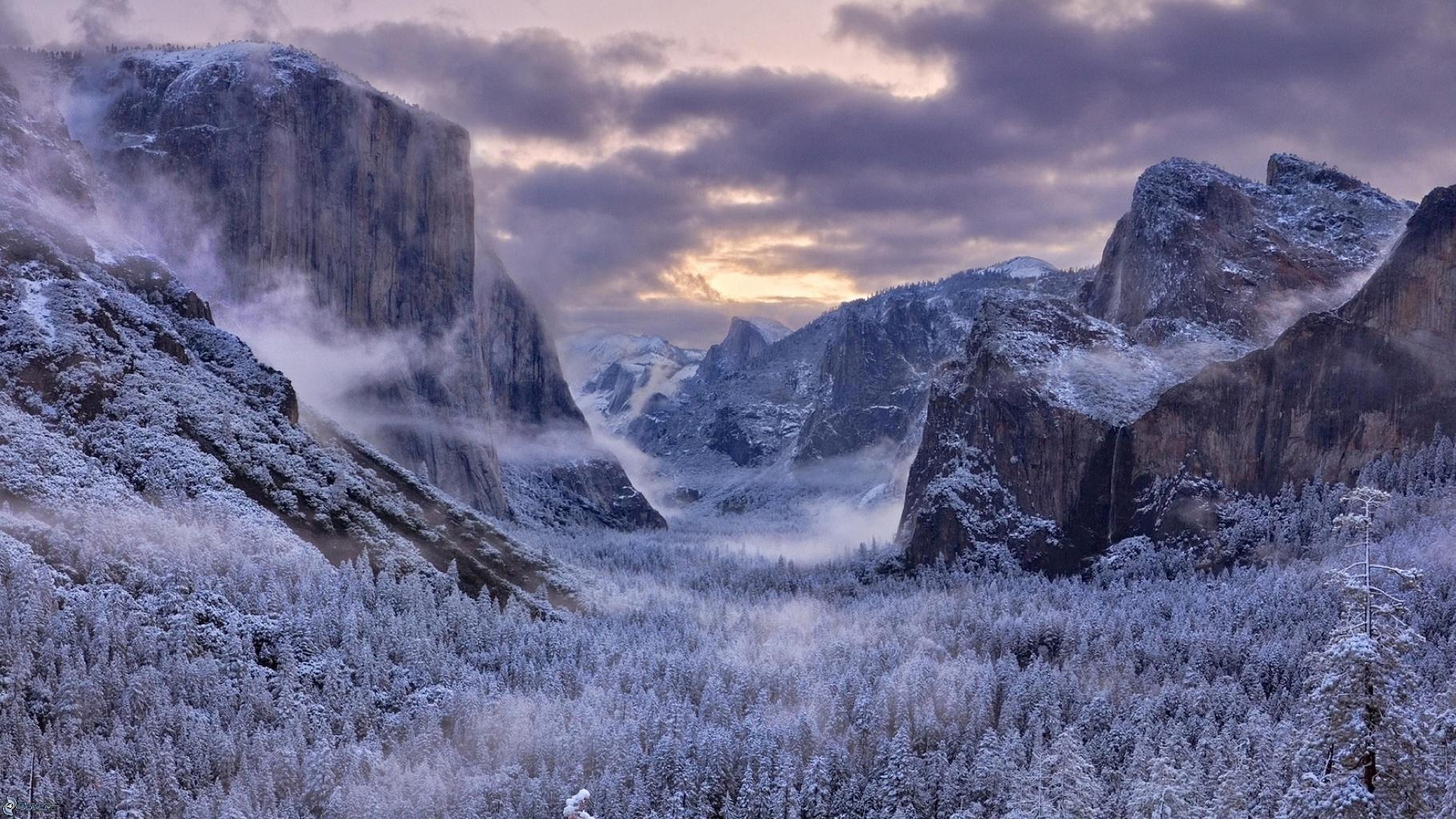} & {\includegraphics[width=\sizea]{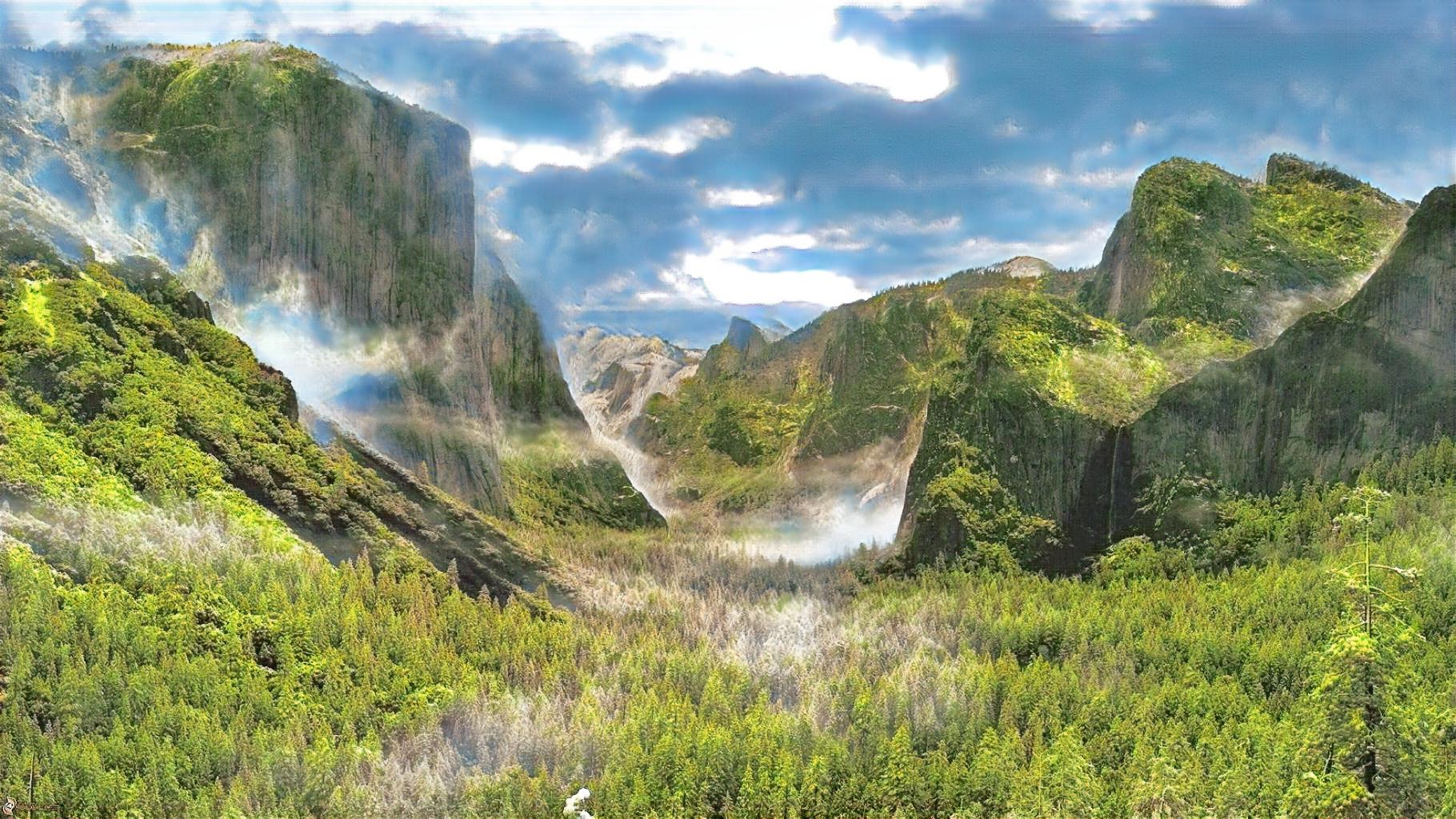}\ \includegraphics[width=\sizea]{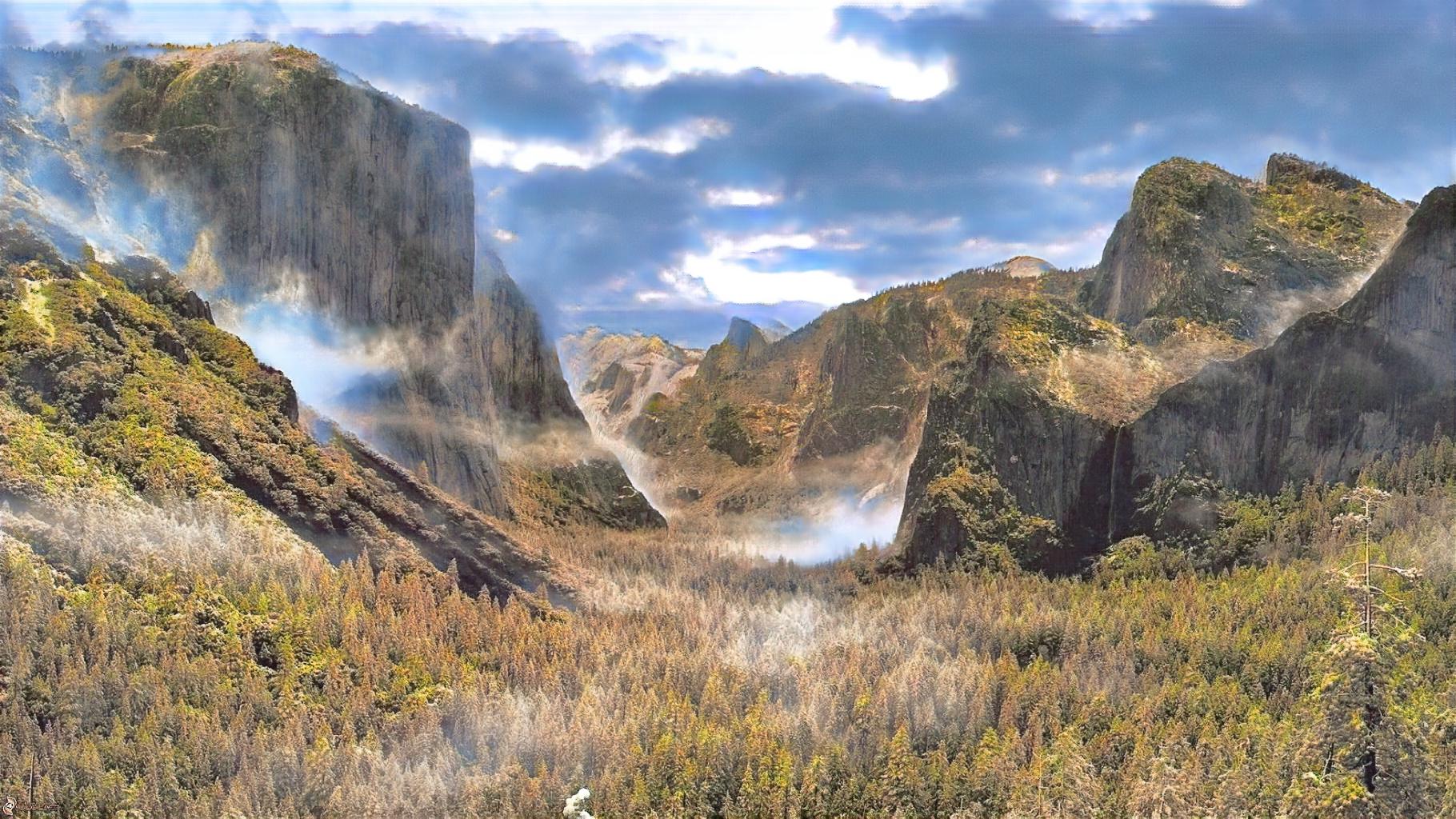}\ \includegraphics[width=\sizea]{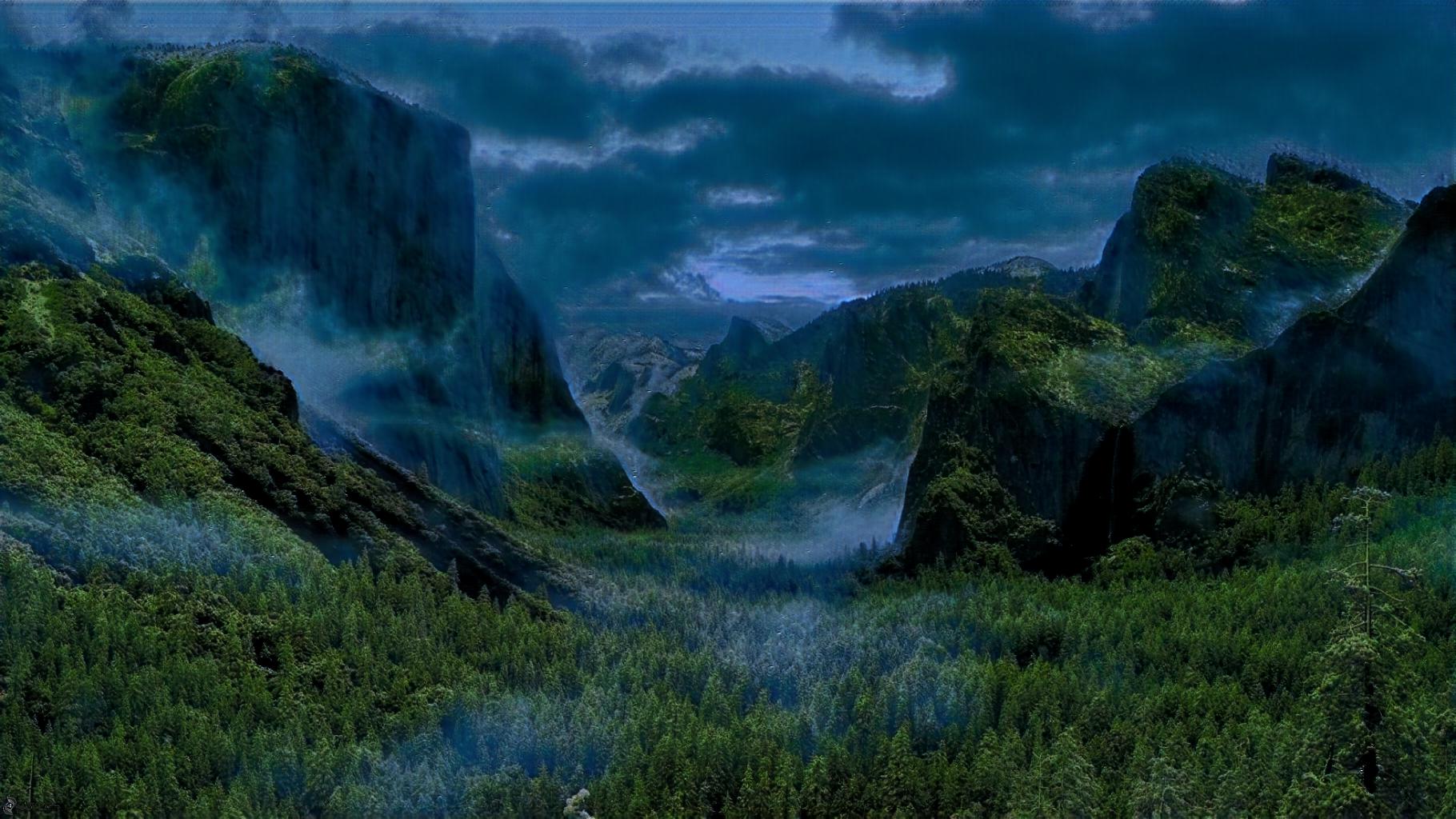}}
		\end{tabular}}{(b) Yosemite winter $\rightarrow$ summer}
		\vspace{-0.1in}	
		\caption{Example results on Yosemite summer $\leftrightarrow$ winter (HD resolution).}
		\label{fig:examples_yosemite}
		\vspace{-0.2in}
	\end{figure*}
	
	\begin{figure*}[!tb]
		\centering
		\small
		\newcommand{\sizea}{0.114\linewidth}	
		\includegraphics[width=\sizea]{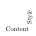}
		\includegraphics[width=\sizea]{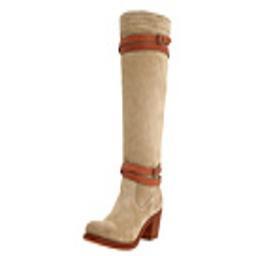}	
		\includegraphics[width=\sizea]{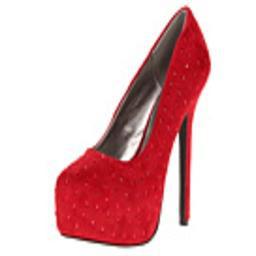}	
		\includegraphics[width=\sizea]{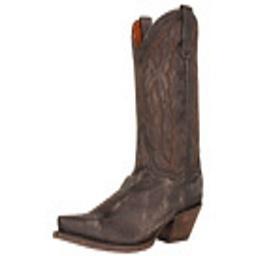}\hspace{0.1in}
		\includegraphics[width=\sizea]{figures/no_sep.pdf}
		\includegraphics[width=\sizea]{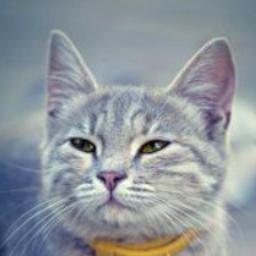}	
		\includegraphics[width=\sizea]{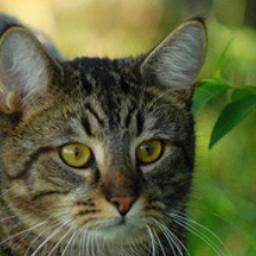}	
		\includegraphics[width=\sizea]{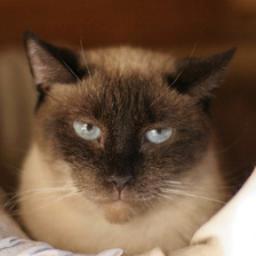}\vspace{0.03in}\\	
		\includegraphics[width=\sizea]{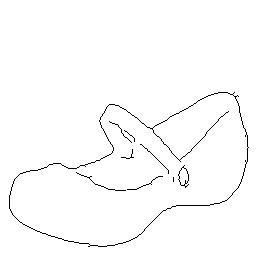}	
		\includegraphics[width=\sizea]{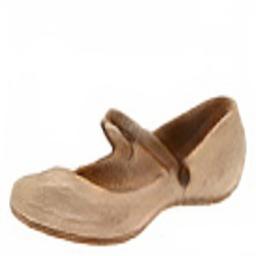}	
		\includegraphics[width=\sizea]{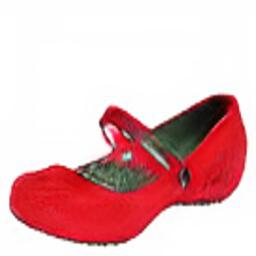}	
		\includegraphics[width=\sizea]{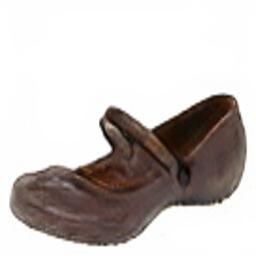}\hspace{0.1in}	
		\includegraphics[width=\sizea]{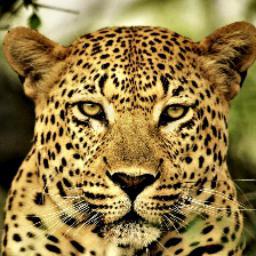}	
		\includegraphics[width=\sizea]{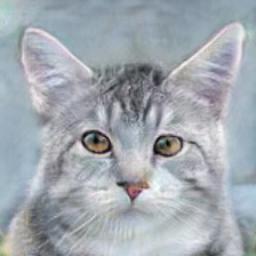}	
		\includegraphics[width=\sizea]{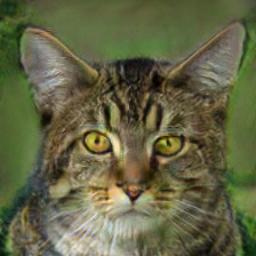}	
		\includegraphics[width=\sizea]{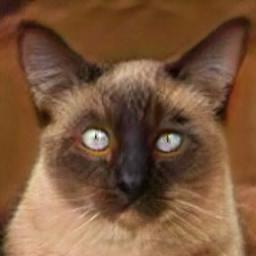}\vspace{0.03in}\\
		\includegraphics[width=\sizea]{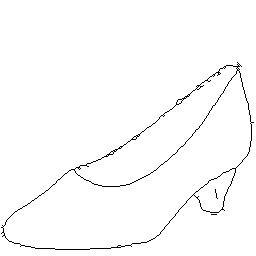}	
		\includegraphics[width=\sizea]{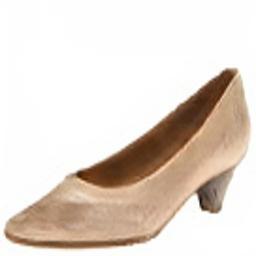}	
		\includegraphics[width=\sizea]{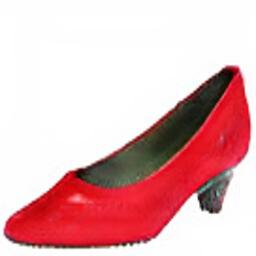}	
		\includegraphics[width=\sizea]{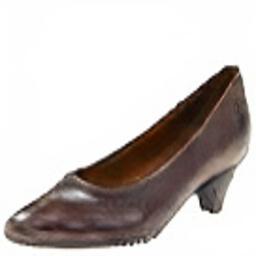}\hspace{0.1in}	
		\includegraphics[width=\sizea]{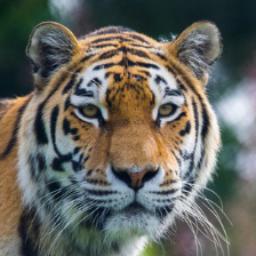}	
		\includegraphics[width=\sizea]{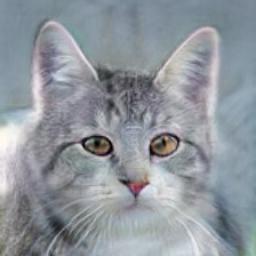}	
		\includegraphics[width=\sizea]{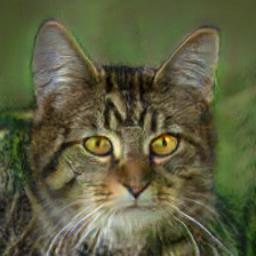}	
		\includegraphics[width=\sizea]{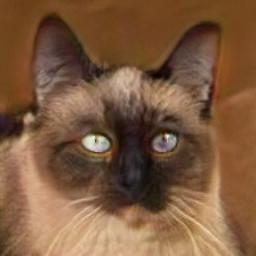}\vspace{0.03in}\\
		\lstackunder[11pt]{\includegraphics[width=\sizea]{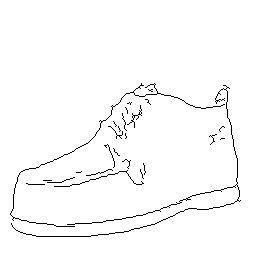}	
			\includegraphics[width=\sizea]{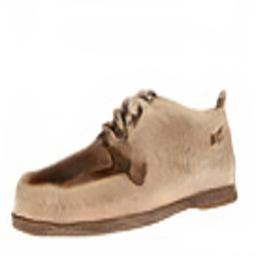}	
			\includegraphics[width=\sizea]{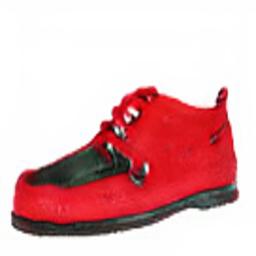}	
			\includegraphics[width=\sizea]{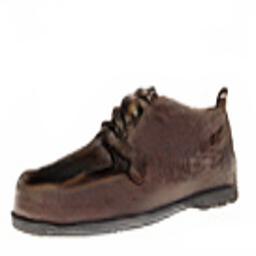}}{(a) edges $\rightarrow$ shoes}\hspace{0.1in}
		\lstackunder[11pt]{\includegraphics[width=\sizea]{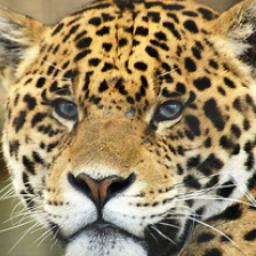}	
			\includegraphics[width=\sizea]{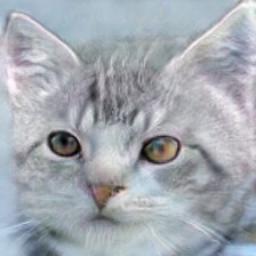}	
			\includegraphics[width=\sizea]{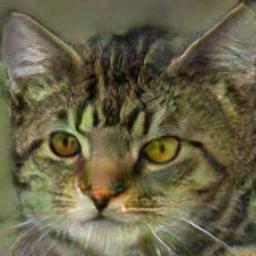}	
			\includegraphics[width=\sizea]{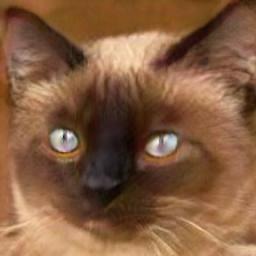}}{(b) big cats $\rightarrow$ house cats}
				\vspace{-0.1cm}
		\caption{ image translation. Each row has the same content while each column has the same style. The color of the generated shoes and the appearance of the generated cats can be specified by providing example style images.}
		\label{fig:style_transfer}
	\end{figure*}
	
	We proceed to perform experiments on the animal image translation dataset. 
	As shown in Fig.~\ref{fig:examples_animals}, our model successfully translate one kind of animal to another. 
	Given an input image, the translation outputs cover multiple modes, \textit{i.e.}, multiple fine-grained animal categories in the target domain.
	The shape of an animal has undergone significant transformations, but the pose is overall preserved. As shown in Table~\ref{tab:cis}, our model obtains the highest scores according to both CIS and IS. In particular, the baselines all obtain a very low CIS, indicating their failure to generate multimodal outputs from a given input. As the IS has been shown to correlate well to image quality~\cite{salimans2016improved}, the higher IS of our method suggests that it also generates images of high quality than baseline approaches.

	Fig.~\ref{fig:examples_street} shows results on street scene datasets. Our model is able to generate SYNTHIA images with diverse renderings (\textit{e.g.}, rainy, snowy, sunset) from a given Cityscape image, and generate Cityscape images with different lighting, shadow, and road textures from a given SYNTHIA image. Similarly, it generates winter images with different amount of snow from a given summer image, and summer images with different amount of leafs from a given winter image. 
	Fig.~\ref{fig:examples_yosemite} shows example results of summer $\leftrightarrow$ winter transfer on the high-resolution Yosemite dataset. Our algorithm generates output images with different lighting.

	\begin{table}[!tb]
		\addtolength{\tabcolsep}{3.3pt}
		\renewcommand\arraystretch{1.2}
		\centering
		\small
		\caption{Quantitative evaluation on animal image translation. This dataset \mbox{contains $3$} domains. We perform bidirectional translation for each domain pair, resulting in $6$ translation tasks. We use CIS and IS to measure the performance on each task. To obtain a high CIS/IS score, a model needs to generate samples that are both high-quality and diverse. While IS measures diversity of all output images, CIS measures diversity of outputs conditioned on a single input image.\label{tab:cis}}
		\vspace{-0.1in}
		\begin{tabular}{|c|c|c|c|c|c|c|c|c|c|}
			\hline
			& \multicolumn{2}{c|}{CycleGAN} & \multicolumn{2}{c|}{\gape{\makecell{CycleGAN* \\ with noise}}} & \multicolumn{2}{c|}{UNIT} & \multicolumn{2}{c|}{MUNIT} \\
			\hline
			& CIS & IS & CIS & IS  & CIS & IS  & CIS & IS    \\ 
			\hline
			house cats $\rightarrow$ big cats    & 0.078 &  0.795  & 0.034 &  0.701  &   0.096   &   0.666  & 0.911 &   0.923  \\ 
			big cats $\rightarrow$ house cats  & 0.109 &  0.887   & 0.124 & 0.848   &  0.164   &   0.817   & 0.956 &   0.954  \\ 
			house cats $\rightarrow$ dogs & 0.044 &  0.895   & 0.070 &  0.901 &   0.045  &   0.827  & 1.231 &   1.255 \\ 
			dogs $\rightarrow$ house cats & 0.121 &  0.921    & 0.137 &  0.978 &   0.193  &   0.982  & 1.035 &   1.034 \\ 
			big cats $\rightarrow$ dogs   & 0.058 &  0.762    & 0.019 &  0.589 &   0.094  &   0.910  & 1.205 &   1.233 \\ 
			dogs $\rightarrow$ big cats   & 0.047 &  0.620    & 0.022 &  0.558 &   0.096  &   0.754  & 0.897 &   0.901 \\ \hline
			Average     & 0.076 &  0.813  & 0.068 &  0.762 &   0.115  &   0.826  & 1.039 &   1.050  \\ 
			\hline
		\end{tabular}
				\vspace{-0.1in}
	\end{table}

\begin{figure*}[!tb]
	\centering
	\small
	\newcommand{\sizea}{0.136\linewidth}	
	\lstackon[60pt]{\includegraphics[width=\sizea]{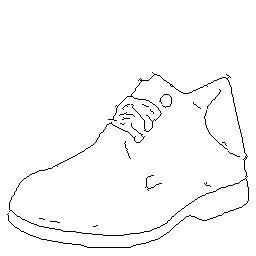}}{{\small Input}}
	\lstackon[60pt]{\includegraphics[width=\sizea]{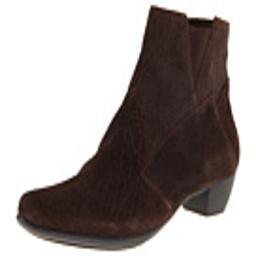}}{Style}
	\lstackon[60pt]{\includegraphics[width=\sizea]{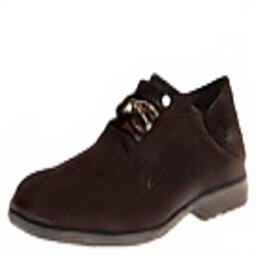}}{Ours}
	\lstackon[60pt]{\includegraphics[width=\sizea]{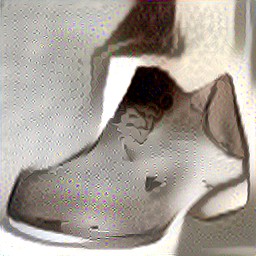}}{Gatys\hspace{0.05cm}\textit{et\hspace{0.1cm}al.}}
	\lstackon[60pt]{\includegraphics[width=\sizea]{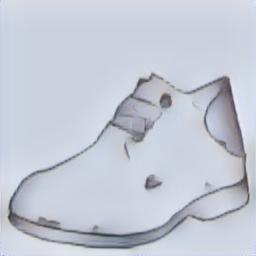}}{Chen\hspace{0.05cm}\textit{et\hspace{0.1cm}al.}}
	\lstackon[60pt]{\includegraphics[width=\sizea]{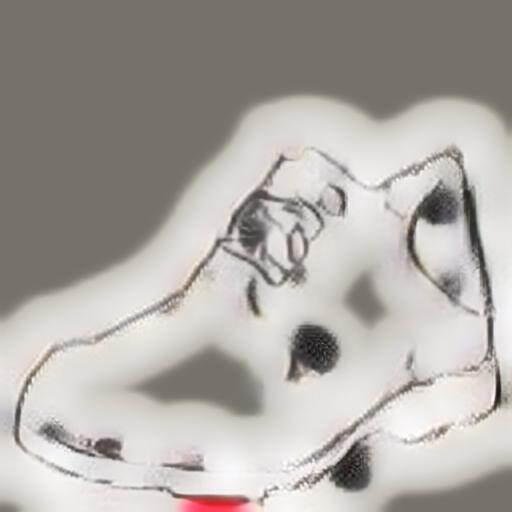}}{AdaIN}
	\lstackon[60pt]{\includegraphics[width=\sizea]{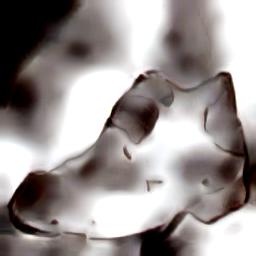}}{WCT}\vspace{0.03in}\\
	\includegraphics[width=\sizea]{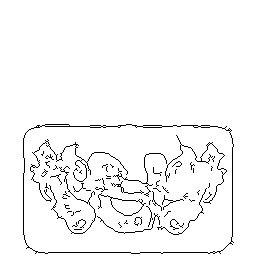}
	\includegraphics[width=\sizea]{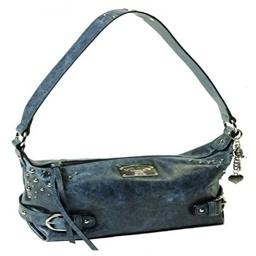}
	\includegraphics[width=\sizea]{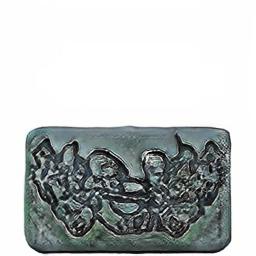}
	\includegraphics[width=\sizea]{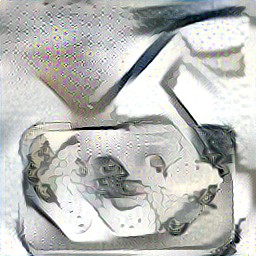}
	\includegraphics[width=\sizea]{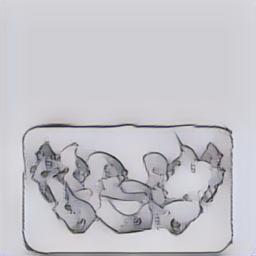}
	\includegraphics[width=\sizea]{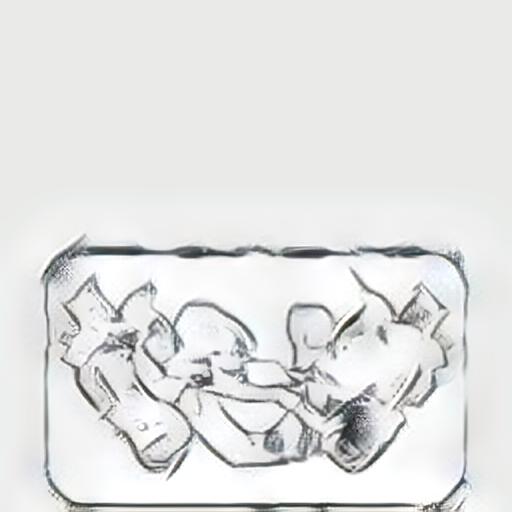}
	\includegraphics[width=\sizea]{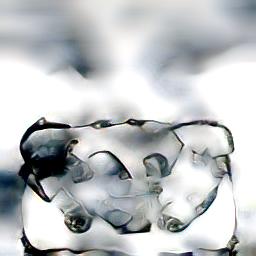}\vspace{0.03in}\\
	\includegraphics[width=\sizea]{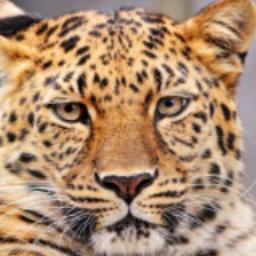}
	\includegraphics[width=\sizea]{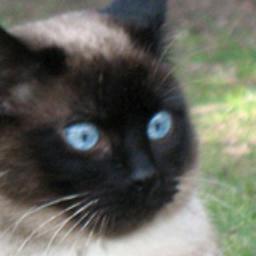}
	\includegraphics[width=\sizea]{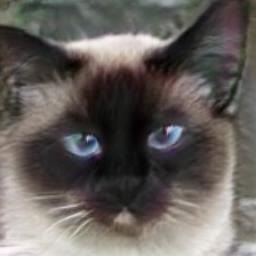}
	\includegraphics[width=\sizea]{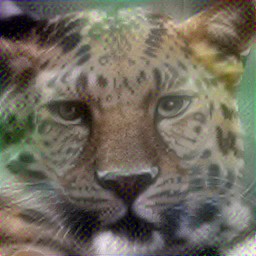}
	\includegraphics[width=\sizea]{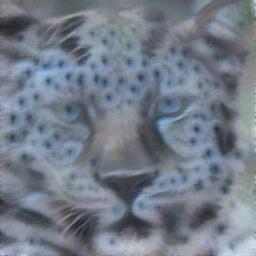}
	\includegraphics[width=\sizea]{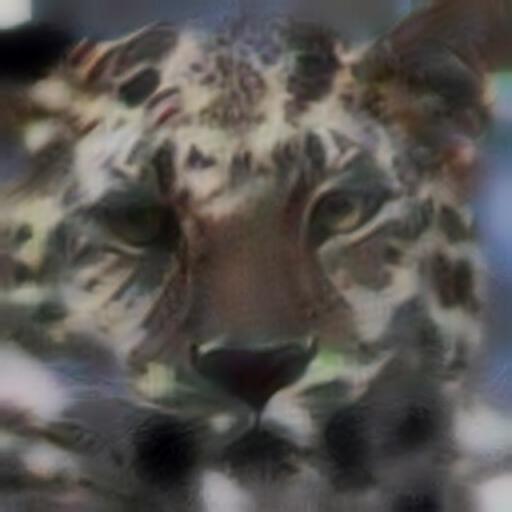}
	\includegraphics[width=\sizea]{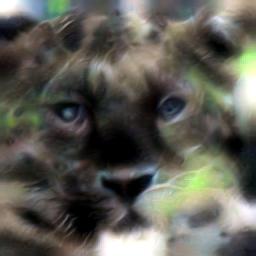}\vspace{0.03in}\\
	\includegraphics[width=\sizea]{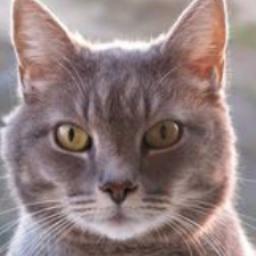}
	\includegraphics[width=\sizea]{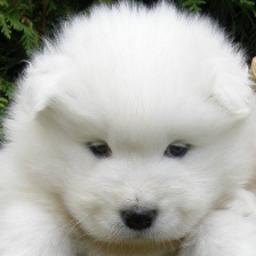}
	\includegraphics[width=\sizea]{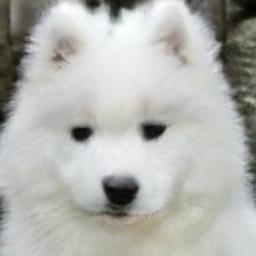}
	\includegraphics[width=\sizea]{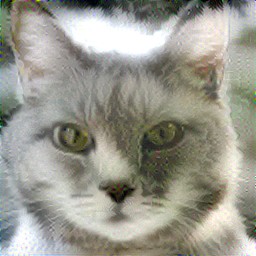}
	\includegraphics[width=\sizea]{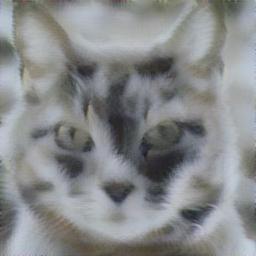}
	\includegraphics[width=\sizea]{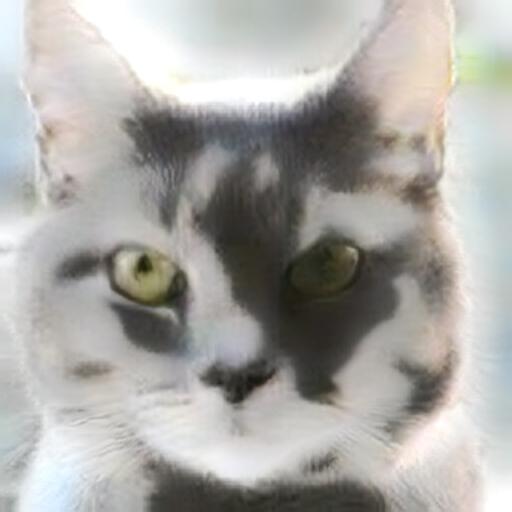}
	\includegraphics[width=\sizea]{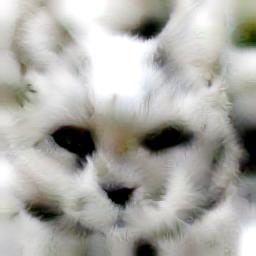}
	\caption{Comparison with existing style transfer methods.}
	\label{fig:style_transfer2}
\end{figure*}

\vpara{Example-guided Image Translation.}  
Instead of sampling the style code from the prior, it is also possible to extract the style code from a reference image. Specifically, given a content image $x_{1}\in\mathcal{X}_{1}$ and a style image $x_{2}\in\mathcal{X}_{2}$, our model produces an image $x_{1\rightarrow 2}$ that recombines the content of the former and the style latter by $x_{1\rightarrow 2} = G_{2}(E^{c}_{1}(x_{1}), E^{s}_{2}(x_{2}))$.  Examples are shown in Fig.~\ref{fig:style_transfer}. 
Note that this is similar to classical style transfer algorithms~\cite{gatys2016image,hertzmann2001image,li2016combining,johnson2016perceptual,huang2017adain,li2017universal,li2018closed} that transfer the style of one image to another. 
In Fig.~\ref{fig:style_transfer2}, we compare out method with classical style transfer algorithms including Gatys~\etal~\cite{gatys2016image}, Chen~\etal~\cite{chen2016fast}, AdaIN~\cite{huang2017adain}, and WCT~\cite{li2017universal}. Our method produces results that are significantly more faithful and realistic, since our method learns the distribution of target domain images using GANs.

	\section{Conclusions}
	We presented a framework for multimodal unsupervised image-to-image translation. Our model achieves quality and diversity superior to existing unsupervised methods and comparable to state-of-the-art supervised approach. Future work includes extending this framework to other domains, such as videos and text.
	
	
	\bibliographystyle{splncs}
	\bibliography{egbib}

	\appendix
	\section{Proofs}
	\label{app:proofs}
	
	\oneshot{1}{
		Suppose there exists $E^{*}_{1}$, $E^{*}_{2}$, $G^{*}_{1}$, $G^{*}_{2}$ such that: 1) $E^{*}_{1} = (G^{*}_{1})^{-1}$ and $E^{*}_{2} = (G^{*}_{2})^{-1}$, and 2) $p(x_{1\rightarrow 2}) = p(x_{2})$ and $p(x_{2\rightarrow 1}) = p(x_{1})$. Then $E^{*}_{1}$, $E^{*}_{2}$, $G^{*}_{1}$, $G^{*}_{2}$ minimizes $\mathcal{L}(E_{1}, E_{2}, G_{1}, G_{2})=\underset{D_{1}, D_{2}}\max\ \mathcal{L}(E_{1}, E_{2}, G_{1}, G_{2}, D_{1}, D_{2})$ (Eq.~(5)).
	}
	\begin{proof}
		\begin{align}
		\mathcal{L}(E_{1}, E_{2}, G_{1}, G_{2})=\ \underset{D_{1}, D_{2}}\max\ \mathcal{L}(E_{1}, E_{2}, G_{1}, G_{2}, D_{1}, D_{2})\notag =\ \underset{D_{1}}\max\ \mathcal{L}^{x_{1}}_{\text{GAN}} + \underset{D_{2}}\max\ \mathcal{L}^{x_{2}}_{\text{GAN}}\\+
		\lambda_{x}(\mathcal{L}^{x_{1}}_{\text{recon}}+\mathcal{L}^{x_{2}}_{\text{recon}})+\notag\ \lambda_{c}(\mathcal{L}^{c_{1}}_{\text{recon}}+\mathcal{L}^{c_{2}}_{\text{recon}})+\lambda_{s}(\mathcal{L}^{s_{1}}_{\text{recon}}+\mathcal{L}^{s_{2}}_{\text{recon}})\notag
		\end{align}
		As shown in Goodfellow~\etal~\cite{goodfellow2014generative}, $\underset{D_{2}}\max\ \mathcal{L}^{x_{2}}_{\text{GAN}}=2\cdot \text{JSD}(p(x_{2})|p(x_{1\rightarrow 2}))-\log4$ which has a global minimum when $p(x_{2})=p(x_{1\rightarrow 2})$. Also, the bidirectional reconstruction loss terms are minimized when $E_{i}$ inverts $G_{i}$. Thus the total loss is minimized under the two stated conditions. Below, we assume the networks have sufficient capacity and the optimality is reachable as in prior works~\cite{goodfellow2014generative,li2017alice}. That is $E_1\rightarrow E_1^*$, 	$E_2\rightarrow E_2^*$, $G_1\rightarrow G_1^*$, and $G_2\rightarrow G_2^*$.
	\end{proof}

\oneshot{2}{\vspace{-0.1in}
	When optimality is reached, we have:
	\label{proposition:2}
	$$p(c_{1})=p(c_{2}),\ p(s_{1})=q(s_{1}),\ p(s_{2})=q(s_{2})$$	
}
	\begin{proof}
		Let $z_{1}$ denote the latent code, which is the concatenation of $c_{1}$ and $s_{1}$. We denote the encoded latent distribution by $p_{E}(z_{1})$, which is defined by $z_{1}=E_{1}(x_{1})$ and $x_{1}$ sampled from the data distribution $p(x_{1})$. We denote the latent distribution at generation time by $p(z_{1})$, which is obtained by $s_{1}\sim q(s_{1})$ and $c_{1}\sim p(c_{2})$. The generated image distribution $p_{G}(x_{1})=p(x_{2\rightarrow 1})$ is defined by $x_{1} = G_{1}(z_{1})$ and $z_{1}$ sampled from $p(z_{1})$. According to the change of variable formula for probability density functions:
		\begin{align}
		p_{G}(x_{1}) &= \lvert\frac{\partial G_{1}^{-1}(x_{1})}{\partial x_{1}} \rvert p(G^{-1}_{1}(x_{1}))\notag\\
		p_{E}(z_{1}) &= \lvert\frac{\partial E_{1}^{-1}(z_{1})}{\partial z_{1}} \rvert p(E_{1}^{-1}(z_{1}))\notag
		\end{align}
		According to Proposition 1, we have $p_{G}(x_{1})=p(x_{1})$ and $E_{1}=G^{-1}_{1}$ when optimality is reached. Thus:
		\begin{align}
		p_{E}(z_{1}) &= \lvert\frac{\partial E_{1}^{-1}(z_{1})}{\partial z_{1}} \rvert p(E_{1}^{-1}(z_{1}))\notag\\
		&= \lvert\frac{\partial E_{1}^{-1}(z_{1})}{\partial z_{1}} \rvert p_{G}(E_{1}^{-1}(z_{1}))\notag\\
		&= \lvert\frac{\partial E_{1}^{-1}(z_{1})}{\partial z_{1}} \rvert  \lvert\frac{\partial G_{1}^{-1}(E_{1}^{-1}(z_{1}))}{\partial E_{1}^{-1}(z_{1})} \rvert	p(G^{-1}_{1}(E_{1}^{-1}(z_{1})))\notag\\
		&= \lvert\frac{\partial E_{1}^{-1}(z_{1})}{\partial z_{1}} \rvert \lvert\frac{\partial G_{1}^{-1}(G_{1}(z_{1}))}{\partial E_{1}^{-1}(z_{1})} \rvert 	p(G^{-1}_{1}(G_{1}(z_{1})))\notag\\
		&= \lvert\frac{\partial E_{1}^{-1}(z_{1})}{\partial z_{1}} \rvert \lvert\frac{\partial z_{1}}{\partial E_{1}^{-1}(z_{1})} \rvert 	p(G^{-1}_{1}(G_{1}(z_{1})))\notag\\
		&= p(z_{1})\notag
		\end{align}
		Similarly we have $p_{E}(z_{2})=p(z_{2})$, which together prove the original proposition. From another perspective, we note that $\mathcal{L}^{c_{2}}_{\text{recon}}, \mathcal{L}^{s_{1}}_{\text{recon}}, \mathcal{L}^{x_{1}}_{\text{GAN}}$ coincide with the objective of a WAE~\cite{tolstikhin2018wasserstein} or AAE~\cite{makhzani2015adversarial} in the latent space, which pushes the encoded latent distribution towards the latent distribution at generation time.
	\end{proof}

\oneshot{3}{
	When optimality is reached,  we have
	$p(x_{1}, x_{1\rightarrow 2}) = p(x_{2\rightarrow 1}, x_{2})$.
}

	\begin{proof}
		For the ease of notation we denote the joint distribution $p(x_{1}, x_{1\rightarrow 2})$ by $	p_{1\rightarrow 2}(x_{1}, x_{2})$ and $p(x_{2\rightarrow 1}, x_{2})$ by $	p_{2\rightarrow 1}(x_{1}, x_{2})$. Both densities are zero when $E_{1}^{c}(x_{1}) \neq E_{2}^{c}(x_{2})$. When $E_{1}^{c}(x_{1}) = E_{2}^{c}(x_{2})$, we also have:
		\begin{align}
		p_{1\rightarrow 2}(x_{1}, x_{2}) &= p_{G}(x_{2}|E^{c}_{1}(x_{1}))p(x_{1})\notag\\
		&= \lvert \frac{\partial E^{s}_2(x_{2})}{x_{2}} \rvert q(E^{s}_{2}(x_{2}))p(x_{1})\notag\\
		&= p(x_{2}|E^{c}_{1}(x_{1}))p_{G}(x_{1})\notag\\
		&= p_{2\rightarrow 1}(x_{1}, x_{2})\notag
		\end{align}
	\end{proof}

\oneshot{4}{
	Denote $h_{1}=(x_{1}, s_{2})\in \mathcal{H}_{1}$ and $h_{2}=(x_{2}, s_{1})\in \mathcal{H}_{2}$. $h_{1}, h_{2}$ are points in the joint spaces of image and style. Our model defines a deterministic mapping $F_{1\rightarrow 2}$ from $\mathcal{H}_{1}$ to $\mathcal{H}_{2}$~(and vice versa) by $F_{1\rightarrow 2}(h_{1}) = F_{1\rightarrow 2}(x_{1}, s_{2})\triangleq(G_{2}(E^{c}_{1}(x_{1}), s_{2}), E^{s}_{1}(x_{1}))$. When optimality is achieved, we have $F_{1\rightarrow 2} = F_{2\rightarrow 1}^{-1}$.
}

	\begin{proof}
		\begin{align}
		F_{2\rightarrow 1}(F_{1\rightarrow 2}(x_{1}, s_{2})) &\triangleq F_{2\rightarrow 1}(G_{2}(E^{c}_{1}(x_{1}), s_{2}), E^{s}_{1}(x_{1}))\\ 
		&\triangleq (G_{1}(E^{c}_{2}(G_{2}(E^{c}_{1}(x_{1}), s_{2})), E^{s}_{1}(x_{1})), E^{s}_{2}(G_{2}(E^{c}_{1}(x_{1}), s_{2})))\\
		&= (G_{1}(E^{c}_{2}(G_{2}(E^{c}_{1}(x_{1}), s_{2})), E^{s}_{1}(x_{1})), s_{2})\\
		&= (G_{1}(E^{c}_{1}(x_{1}), E^{s}_{1}(x_{1})), s_{2})\\
		&= (x_{1}, s_{2})
		\end{align}
		And we can prove $F_{1\rightarrow 2}(F_{2\rightarrow 1}(x_{2}, s_{1})) = (x_{2}, s_{1})$ in a similar manner. To be more specific, $(3)$ is implied by the style reconstruction loss $\mathcal{L}^{s}_{\text{recon}}$, $(4)$ is implied by the content reconstruction loss $\mathcal{L}^{c}_{\text{recon}}$, and $(5)$ is implied by the image reconstruction loss $\mathcal{L}^{x}_{\text{recon}}$. As a result, style-augmented cycle consistency is implicitly implied by the proposed bidirectional reconstruction loss.
	\end{proof}

\noindent \textbf{Proposition 5. (Cycle consistency implies deterministic translations).}
	Let $p(x_{1})$ and $p(x_{2})$ denote the data distributions. $p_{G}(x_{1}|x_{2})$ and $p_{G}(x_{2}|x_{1})$ are two conditionals defined by generators. Given 1) matched marginals: $p(x_{1})=\int p_{G}(x_{1}|x_{2})p(x_{2})\,dx_{2}$, $p(x_{1})=\int p_{G}(x_{2}|x_{1})p(x_{1})\,dx_{1}$, and 2) cycle consistency: $\mathbb{E}_{x_{2}\sim p_{G}(x_{2}|x^{*}_{1})}[p_{G}(x_{1}|x_{2})]=\delta(x_{1}-x^{*}_{1})$, $\mathbb{E}_{x_{1}\sim p_{G}(x_{1}|x^{*}_{2})}[p_{G}(x_{2}|x_{1})]$$=\delta(x_{2}-x^{*}_{2})$ for every $x^{*}_{1}\in\mathcal{X}_{1}, x^{*}_{2}\in\mathcal{X}_{2}$, then $p_{G}(x_{1}|x_{2})$ and $p_{G}(x_{2}|x_{1})$ collapse to deterministic delta functions.

	\begin{proof}
		Let $x^{*}_{1}$ be a sample from $p(x_{1})$. $x^{\prime}_{2}$, $x^{\prime\prime}_{2}$ are two samples from $p_{G}(x_{2}|x^{*}_{1})$. Due to cycle consistency in $\mathcal{X}_{1}\rightarrow \mathcal{X}_{2}\rightarrow \mathcal{X}_{1}$, we have $p_{G}(x_{1}|x^{\prime}_{2})=p_{G}(x_{1}|x^{\prime\prime}_{2})=\delta(x_{1}-x^{*}_{1})$. Also, $x^{\prime}_{2}\in\mathcal{X}_{2}$ and $x^{\prime\prime}_{2}\in \mathcal{X}_{2}$ because of matched marginals. Due to cycle consistency in $\mathcal{X}_{2}\rightarrow \mathcal{X}_{1}\rightarrow \mathcal{X}_{2}$, we have $p_{G}(x_{2}|x^{*}_{1})=\delta(x_{2}-x^{\prime}_{2})=\delta(x_{2}-x^{\prime\prime}_{2})$. Thus $p_{G}(x_{2}|x_{1})$ collapses to a delta function, similar for $p_{G}(x_{1}|x_{2})$.		
		This proposition shows that cycle consistency~\cite{zhu2017unpaired} is a too strong constraint for multimodal image translation.
	\end{proof}
	
	\section{Training Details}
	\label{app:hyperparameters}
	\subsection{Hyperparameters}
	We use the Adam optimizer~\cite{kingma2015adam} with $\beta_{1}=0.5$, $\beta_{2}=0.999$, and an initial learning rate of $0.0001$. The learning rate is decreased by half every $100,000$ iterations. In all experiments, we use a batch size of 1 and set the loss weights to $\lambda_{x}=10$, $\lambda_{c}=1$, $\lambda_{s}=1$. We use the domain-invariant perceptual loss with weight $1$ in the street scene and Yosemite datasets. We choose the dimension of the style code to be $8$ across all datasets. Random mirroring is applied during training.
	\subsection{Network Architectures}
	Let \texttt{c7s1-k} denote a $7\times 7$ convolutional block with k filters and stride 1. \texttt{dk} denotes a $4\times4$ convolutional block with k filters and stride 2. \texttt{Rk} denotes a residual block that contains two $3\times3$ convolutional blocks. \texttt{uk} denotes a $2\times$ nearest-neighbor upsampling layer followed by a $5 \times 5$ convolutional block with k filters and stride 1. \texttt{GAP} denotes a global average pooling layer. \texttt{fck} denotes a fully connected layer with k filters. We apply Instance Normalization~(IN)~\cite{ulyanov2017improved} to the content encoder and Adaptive Instance Normalization~(AdaIN)~\cite{huang2017adain} to the decoder. We use ReLU activations in the generator and Leaky ReLU with slope $0.2$ in the discriminator. We use multi-scale discriminators with $3$ scales.
	\begin{itemize}
		\item Generator architecture
		\begin{itemize} 
			\item Content encoder: \texttt{c7s1-64, d128, d256, R256, R256, R256, R256}
			\item Style encoder: \texttt{c7s1-64, d128, d256, d256, d256, GAP, fc8}
			\item Decoder: \texttt{R256, R256, R256, R256, u128, u64, c7s1-3}
		\end{itemize}
		\item Discriminator architecture: \texttt{d64, d128, d256, d512}
	\end{itemize}
	
	\section{Domain-invariant Perceptual Loss}
	\label{app:perceptual}

\begin{figure*}[!t]
		\centering
		\small
		\newcommand{\sizea}{0.23\linewidth}	
		\lstackunder[14pt]{\includegraphics[width=\sizea]{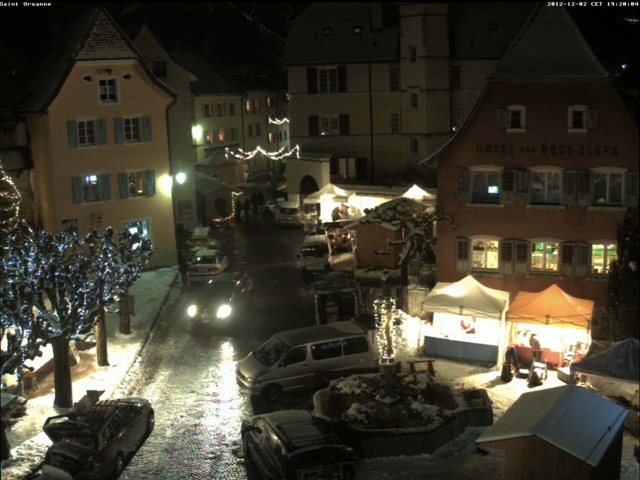}\ \includegraphics[width=\sizea]{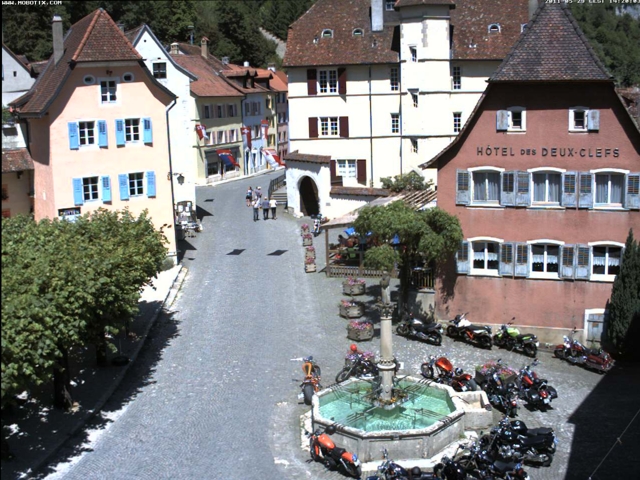}}{(a) Image pairs from the same scene.}\hspace{0.1in}
		\lstackunder[14pt]{\includegraphics[width=\sizea]{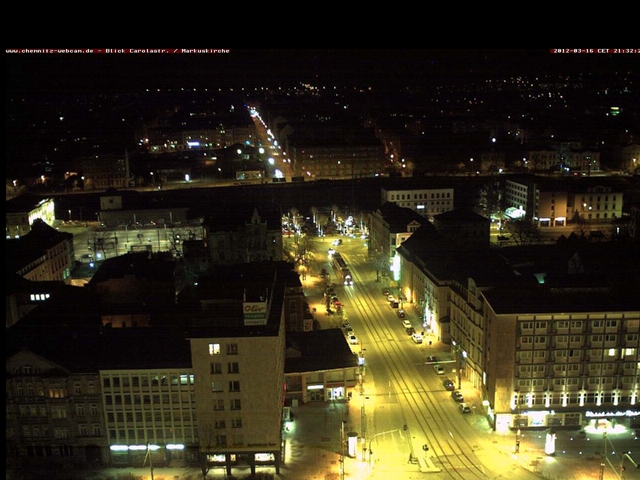}\ \includegraphics[width=\sizea]{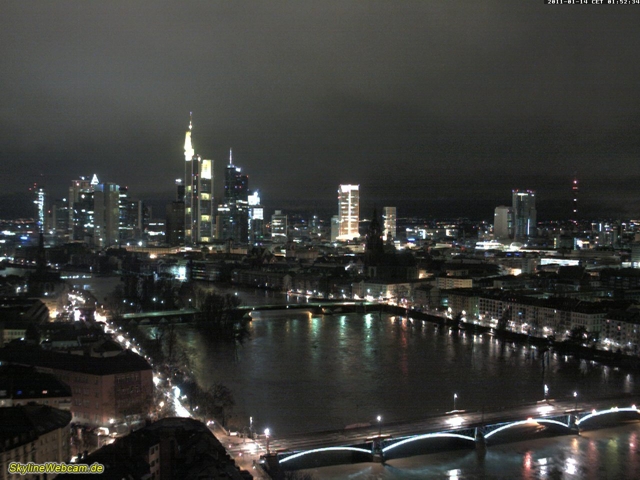}}{(b) Image pairs from the same domain.}
		\caption{Example image pairs for domain-invariant perceptual loss experiments.}
		\label{fig:examples}
	\end{figure*}

\begin{figure*}[!t]
		\centering
		\small
		\newcommand{\sizea}{0.49\linewidth}	
		\includegraphics[width=\sizea]{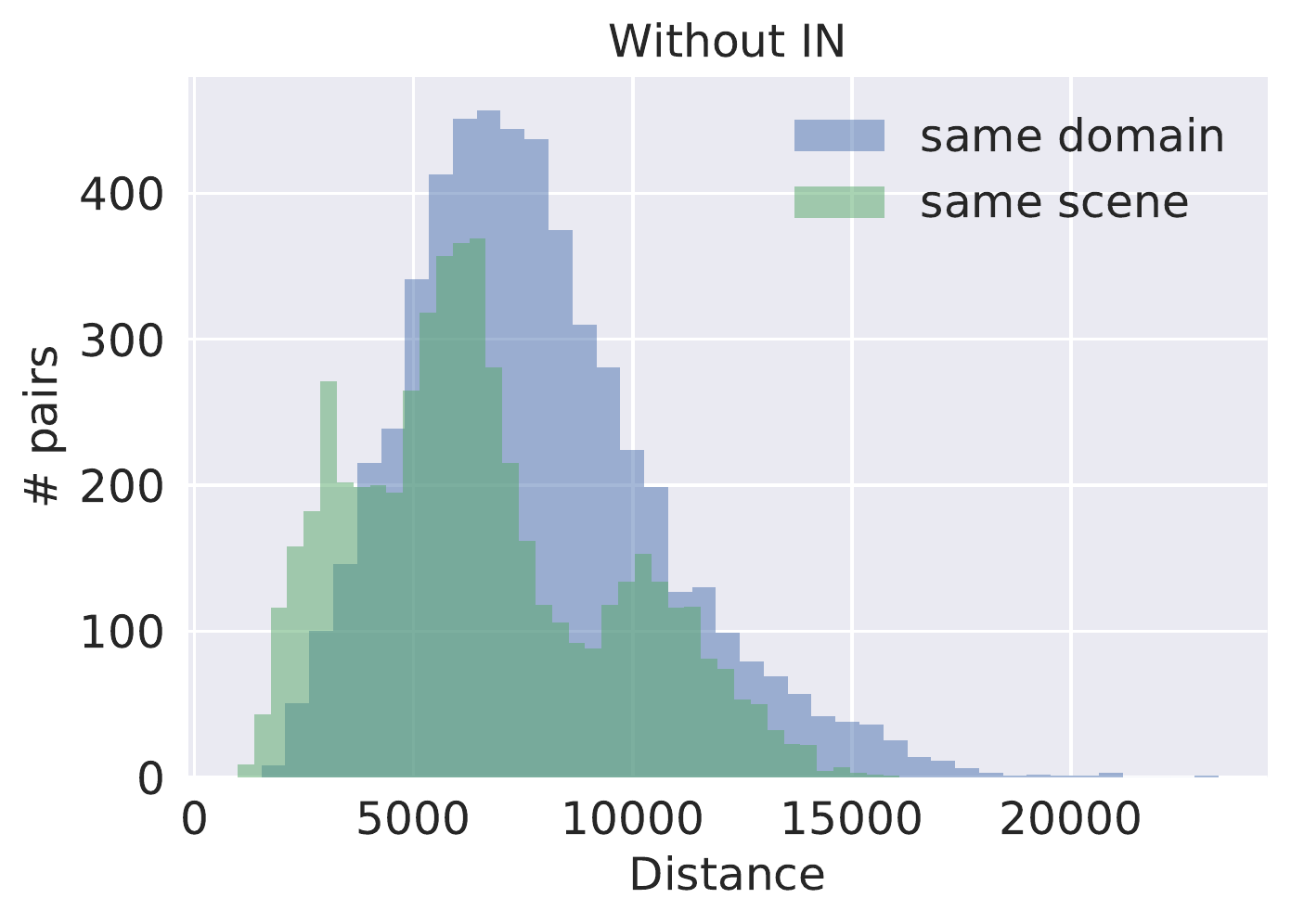}
		\includegraphics[width=\sizea]{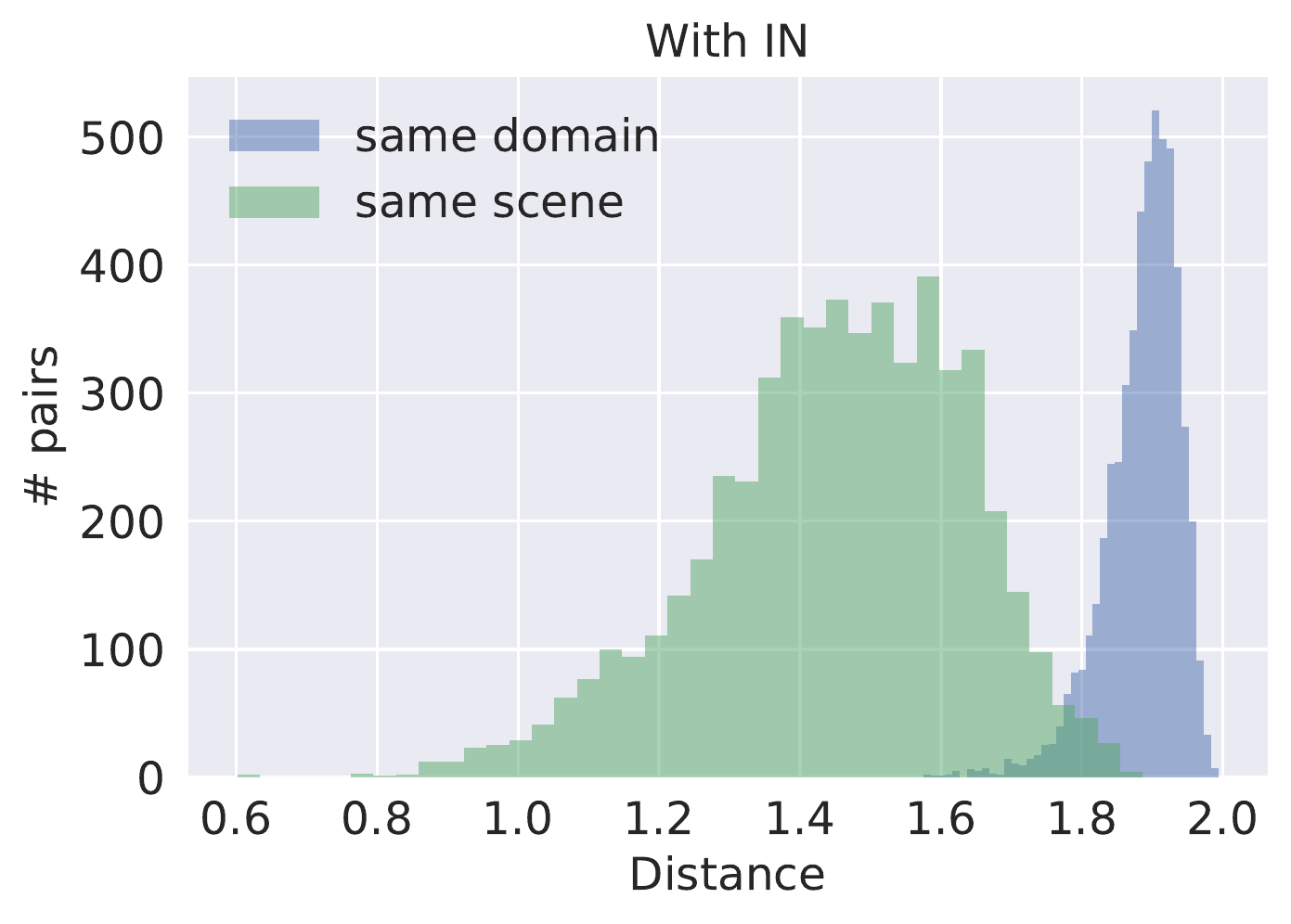}
		\caption{Histograms of the VGG feature distance. Left: distance computed without using IN. Right: distance computed after IN. Blue: distance between image pairs from the same domain (but different scenes). Green: distance between image pairs from the same scene~(but different domains).}
		\label{fig:vgg_loss}
	\end{figure*}

	We conduct an experiment to verify if applying IN before computing the feature distance can indeed make the distance more domain-invariant. We experiment on the day $\leftrightarrow$ dataset used by Isola~\etal~\cite{isola2017image} and originally proposed by Laffont~\etal~\cite{laffont2014transient}. 
	We randomly sample two sets of image pairs: 1) images from the same domain~(both day or both night) but different scenes, 2) images from the same scene but different domains. 
	Fig.~\ref{fig:examples} shows examples from the two sets of image pairs.
	We then compute the VGG feature~(\texttt{relu4\_3}) distance between each image pair, with IN either applied or not before computing the distance. In Fig.~\ref{fig:vgg_loss}, we show histograms of the distance computed either with or without IN, and from image pairs either of the same domain or the same scene. Without applying IN before computing the distance, the distribution of feature distance is similar for both sets of image pairs. With IN enabled, however, image pairs from the same scene have clearly smaller distance, even they come from different domains. The results suggest that applying IN before computing the distance makes the feature distance much more domain-invariant.
	
\end{document}